\documentclass[10pt,twocolumn,letterpaper]{article}

\usepackage[pagenumbers]{cvpr} 

\usepackage[accsupp]{axessibility}  

\definecolor{cvprblue}{rgb}{0.21,0.49,0.74}
\usepackage[pagebackref,breaklinks,colorlinks,allcolors=cvprblue]{hyperref}
\usepackage{color,soul}

\usepackage{algorithm}
\usepackage{algpseudocode}
\usepackage{placeins}

\def\paperID{xxxxx} 
\def\confName{CVPR}
\def\confYear{2026}

\title{JACoP: Joint Alignment for Compliant Multi-Agent Prediction}

\author{
Qingze (Tony) Liu$^{1}$ \quad
Alen Mrdovic$^{1}$ \quad
Danrui Li$^{1}$ \quad
Mathew Schwartz$^{1}$ \\
Sejong Yoon$^{2}$\quad
Mubbasir Kapadia$^{1}$ \\
$^{1}$Rutgers University, New Brunswick \\
$^{2}$The College of New Jersey
}

\begin{document}
\maketitle
\begin{abstract}
Stochastic Human Trajectory Prediction (HTP) using generative modeling has emerged as a significant area of research. Although state-of-the-art models excel in optimizing the accuracy of individual agents, they often struggle to generate predictions that are collectively compliant, leading to output trajectories marred by social collisions and environmental violations, thus rendering them impractical for real-world applications. To bridge this gap, we present JACoP: Joint Alignment for Compliant Multi-Agent Prediction, an innovative multi-stage framework that ensures scene-level plausibility. JACoP incorporates an Anchor-Based Agent-Centric Profiler for effective initial compliance filtering and employs a Markov Random Field (MRF) based aligner to formalize the joint selection for scene predictions. By representing inter-agent spatial and social costs as MRF energy potentials, we successfully infer and sample from the joint trajectory distribution, achieving prediction with optimal scene compliance. Comprehensive experiments show that JACoP not only achieves competitive accuracy, but also sets a new standard in reducing both environmental violations and social collisions, thereby confirming its ability to produce collectively feasible and practically applicable trajectory predictions. \footnote{\href{https://github.com/TL-QZ/JACoP}{https://github.com/TL-QZ/JACoP}}
\end{abstract}
    
\section{Introduction}
\label{sec:intro}

\begin{figure}
    \centering
    \includegraphics[width=\linewidth]{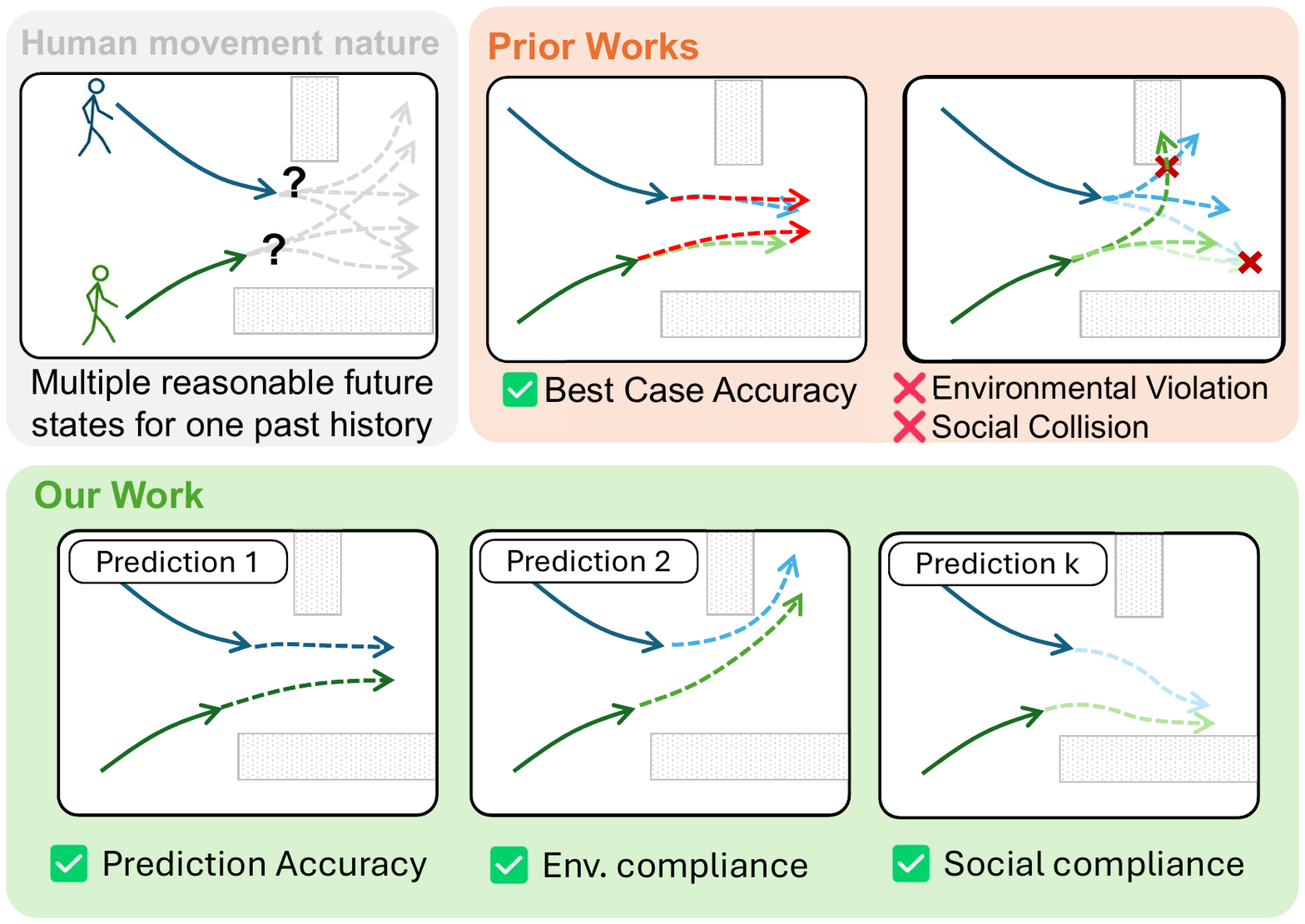}
    \vspace{-0.3in}
    \caption{HTP models need environmentally and socially compliant predictions to maximize their utility in downstream tasks. JACoP produce compliant predictions to accomplish this goal.}
    \label{fig1}
    \vspace*{-0.5cm}
\end{figure}

Human Trajectory Prediction (HTP) studies human behavior at a microscopic level by predicting individual movement patterns given historical observations. Due to the complexity of human decision making, along with the flexibility of pedestrian movements, the HTP problem is stochastic by nature, i.e., a given historical observation might be compatible with multiple feasible future outcomes. The rapid development of generative modeling and its successful application in image generation inspired their adoption in the HTP domain. 

Previous works\cite{fu2025moflow,bae2024singulartrajectory,mao2023leapfrog}, leveraging the strong learning capability of SOTA generative models, focused on optimizing prediction accuracy among its K-shot generative outputs. The addition of social and environmental context as input and the dedicated component for modeling interactions further enhance the model's accuracy upper-bound, measured by the displacement of the best prediction among all model outputs (i.e., $min_kADE/FDE$). To further ensure coverage of a broader range of possible future scenarios and a higher likelihood of recreating ground-truth behavior, previous models also focused on enhancing the diversity of their predictions, either via adoption of alternative learning objective \cite{yuan2021agentformer} or post-hoc sampling modules \cite{mangalam2021goals}. Despite the numerical improvement in the accuracy metric, Our experiment and previsous studies \cite{lee2022muse,liu2024trajdiffuse} have shown that SOTA models often struggle to generate plausible trajectory predictions free of environmental violation and social collision, rendering the majority generated trajectory predictions unusable as reference for downstream planning tasks. \Cref{fig1} depicts our motivation.

To address this issue, we propose Joint Alignment for Compliant Prediction (JACoP), a multi-stage framework designed to enforce scene-level compliance across multiple agents simultaneously. The design of JACoP integrates two key elements. First, we introduce an anchor-based agent-centric profiler, which effectively curates a set of high-quality trajectory prototypes. These prototypes are selected not only to capture a diverse set of agent intentions but also to maintain initial compliance with environmental constraints. This process efficiently constrains the solution space for the subsequent joint sampling stage required for optimal alignment of the scene prediction. Second, we formalize the selection of the final prediction output by inferring a joint trajectory distribution over the selected prototypes through a Markov Random Field (MRF). The MRF explicitly models the spatial and social costs, including joint occurrence likelihood and inter-agent collision penalties, as energy potentials. This allows for the correspondence of the optimal joint trajectory prediction with the lowest-energy configurations of individual prototypes. This mechanism, combined with Gibbs Sampling, effectively models the joint distribution of future trajectories for all scene agents and offers a robust strategy to ensure that the predicted outcomes maintain environmental and social plausibility across the entire scene.

We conduct thorough benchmarking on widely-used trajectory prediction datasets. Additionally, our assessment extends beyond mere accuracy measures of models' best predictions, focusing instead on evaluating their true ability to comprehend scene contexts and simultaneously generate joint scene predictions for all agents. We demonstrate that our approach not only performs competitively in terms of joint accuracy metrics but also sets new benchmarks in reducing environmental violations and social collisions. This confirms that our model effectively closes the gap between precise individual forecasts and the collective feasibility needed for practical human trajectory prediction.

In summary, our contributions are summarized as follows: 
\begin{itemize}
    \item We propose JACoP, a multi-stage trajectory prediction pipeline designed to enforce scene-level compliance across multiple interacting agents simultaneously, bridging the gap between high individual accuracy and collective feasibility.
    \item We formalize the joint selection of the final predictions by inferring a joint trajectory distribution using an MRF. This model explicitly incorporates spatial and social costs, such as inter-agent collision penalties and joint occurrence likelihood, as energy potentials, guaranteeing low-energy (plausible) configurations.
    \item We establish new state-of-the-art performance in minimizing crucial metrics for environmental and social compliance. Our work confirms that JACoP successfully generates joint scene predictions that are not only accurate but also consistently practical and usable for downstream planning tasks.
\end{itemize}

\section{Related Works}
\label{sec:related}

Human Trajectory Prediction has been a prominent topic in the field of computer vision. Early learning-based methods use recurrent neural networks \cite{alahi2016social, gupta2018social} to model the sequential nature of human trajectory. Later works shift toward using generative models to generate multi-modal predictions better capturing the stochastic nature of human trajectory, using techniques such as Generative Adversarial Network (GAN) \cite{gupta2018social, huang2019stgat, kosaraju2019social, moder2021coloss}, Conditional Variational Autoencoders (CVAE) \cite{salzmann2020trajectron++,lee2017desire, lee2022muse, xu2022socialvae}, Normalizing flow \cite{rhinehart2018r2p2, rhinehart2019precog, chen2024mgf}, diffusion model \cite{gu2022stochastic, mao2023leapfrog, liu2024trajdiffuse, capellera2025unified} and Flow Matching Model \cite{fu2025moflow}.

Several studies developed sampling heuristics to improve prediction diversity. AgentFormer~\cite{yuan2021agentformer} uses a sampling module to enhance diversity from the CVAE module's latent distribution; MemoNet~\cite{xu2022remember} generates a large amount of trajectory samples, performs clustering, and uses cluster centers for maximum diversity, and FEND~\cite{wang2023fend} and AMD \cite{rao2025amd} focus on sampling long-tail events for better coverage of possible future scenarios.

Recent studies incorporate environmental factors to improve prediction accuracy through scene contexts. Some~\cite{mangalam2021goals,lee2022muse} use rasterized environmental maps to predict waypoints, whereas others~\cite{liang2020learning, gao2020vectornet,zhou2022hivt,zhou2023query, shi2024mtr++} use HD maps for vehicle trajectory predictions. Unlike vehicles, humans do not need to strictly follow lane structures or non-drivable areas, therefore lacking direct environmental guidance for accurate predictions. To address this issue, methods including ~\cite{varadarajan2021d, bae2024singulartrajectory} used anchor trajectories as initial proposals, which we also adopted in our model.

Social interaction modeling in the HTP domain is more explored than environmental factors. Research has focused on using social contexts to improve K-shot multi-modal prediction accuracy. Initial studies highlighted social feature extraction, introducing social-pooling to merge information from nearby agents into a target agent's embedding \cite{alahi2016social, gupta2018social}. Later, specialized interaction modules like social graphs, using graph-based learning such as GCN~\cite{mohamed2020social, sun2020recursive} and GAT~\cite{huang2019stgat, kosaraju2019social, zhou2023query}, were developed for multiagent contexts. Other methods \cite{bae2022learning} furthered this by adding social graph construction as an auxiliary task and using higher-order graphs \cite{kim2024higher,chen2025socialmoif} to better represent group behaviors of pedestrian agents.

Earlier methods use multi-agent contexts for decoding, assuming independent future trajectories for each agent once conditioned on past social contexts. AgentFormer and FJMP~\cite{yuan2021agentformer,Rowe2023-mt} address this by inferring from the joint distribution for scene consistency. AgentFormer~\cite{yuan2021agentformer} utilizes autoregressive CVAE to model joint agent motion, while FJMP~\cite{Rowe2023-mt} applies topological sorting and conditional sampling. Our work follows this approach, producing scene-consistent multi-agent predictions by sampling from a joint distribution represented by the MRF. We further enhance this by optimizing beyond prediction accuracy to ensure the model understands social contexts, achieving both environmental and social compliance.

Previous studies employed $min_kADE$ to assess model predictions by focusing on the optimal individual agent output. To improve evaluations involving multiple interacting agents, \cite{weng2023joint} presented Joint$ADE$ and Joint$FDE$ (JADE/JFDE) for predicting pedestrian trajectories. While JADE/JFDE provides a more comprehensive joint performance evaluation, it still only assesses the best sample, lacking a full assessment of trajectory generation quality. Inspired by crowd simulation evaluation metrics, \cite{sohn2021a2x} suggested adding qualitative measures like environmental and social collision and diversity metrics to HTP evaluations. These measures assess the overall model capability by considering all multi-modal outputs. In line with this, we incorporate metrics for scene compliance and replicate SOTA models to address this gap in current HTP research.



\section{Method}
\label{sec:method}
\begin{figure*}[t]
    \centering
    \vspace{-0.2in}
    \includegraphics[width=0.99\linewidth]{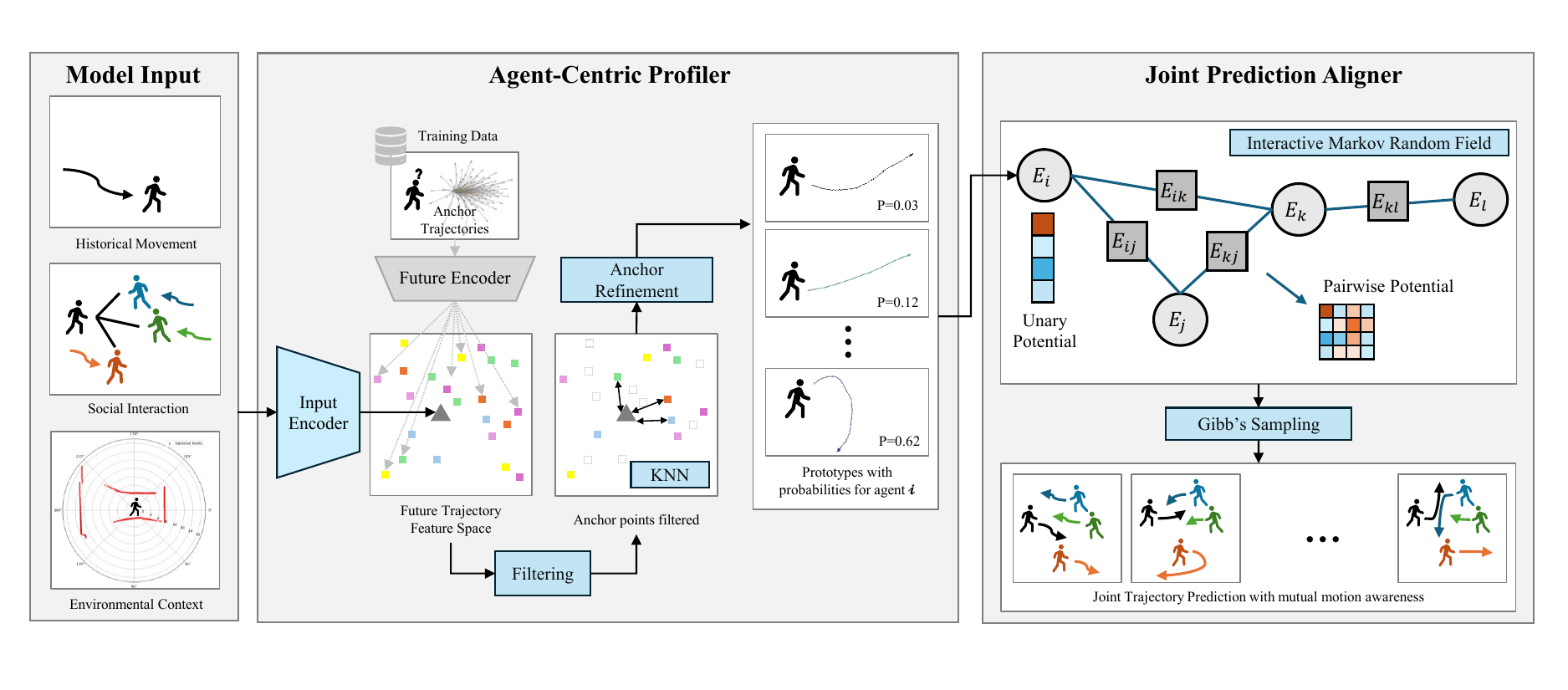}
    \vspace{-0.3in}
\caption{\textbf{Model Architecture.} Our framework operates in two stages: (\textit{Left}) Latent embeddings from agents' historical movement, social context, and environment query prototype trajectories, which are filtered and refined. (\textit{Right}) We then use the refined proposals to infer a joint distribution of future trajectories via a Markov Random Field (MRF), with the final scene prediction sampled using Gibbs sampling.}
    \label{fig:main}
\vspace{-0.2in}
\end{figure*}
\subsection{Overview}
\label{sub:overview}
Human trajectory prediction (HTP) aims to predict the future movement given a sequence of observed behavior and environmental context. Suppose that there exist a total of $N$ agents in a scene and an observed trajectory sequence $X_i=\{x_i^t|t\in[1,T_o]\}$ for each agent $i$ for $T_o$ steps in 2D global coordinates, where $X=\{X_i|i\in[1,N]\}$. We denote the ground-truth (GT) future trajectory for each agent $i$ as $Y_i=\{y_i^t|t\in[T_o+1,T_o+T_f]\}$ for $T_{f}$ steps into the future, and $Y=\{Y_i|i\in[1,N]\}$. Our goal is to jointly predict future trajectories $\hat{Y}$ for all $N$ agents based on historical trajectories $X$ and the environmental context $M$.

\subsection{Agent-Centric Profiler}
\label{sub:profiler}
The Agent-Centric Profiler (ACP) determines the action profile of the target agent through their own historical movement, neighboring agent positions, and environment layout. We use these historical and scene contexts to query against a set of anchor trajectories $\mathbf{Y^*}\subset \mathbb{R}^{M \times T_f \times 2}$ and select $K$ possible future movements as prototype trajectories, producing scores rating for their likelihood in the meantime. 

\noindent
\textbf{Anchor Trajectory Database } We first construct a database of $M$ anchor trajectories from the training set to cover various possible movement types. Inspired by SingularTrajectory~\cite{bae2024singulartrajectory}, we use singular value decomposition (SVD) to compress the GT trajectories $Y_i$ into a lower-dimensional vector $v_i \in \mathbb{R}^{d_s}, d_s=4$, following the previous work. To do so, we first construct a motion matrix $A \in \mathbb{R}^{N\times 2T_f}$ by flattening and concatenating all the normalized ground-truth future trajectories in the training data. Then we use SVD to decompose the matrix $A$ into
\begin{equation}
    A = U\Sigma V
\end{equation}
where $U\in\mathbb{R}^{N\times d_s}$, $\Sigma\in\mathbb{R}^{d_s\times d_s}$ and $V\in\mathbb{R}^{d_s\times 2T_f}$. We use $V^\top$ to obtain the trajectory compression by
\begin{equation}
    v_i = A_{i}V^\top,
\end{equation}
where $A_i\in \mathbb{R}^{2T_f}$ is the normalized and flattened version of $Y_i$ for agent $i$. We then use K-means clustering to group all compressed trajectories into $\mathcal{K}$ clusters and use the cluster centers as the anchor trajectory by transforming them back into the coordinate space to build the database $\mathbf{Y^*}\subset \mathbb{R}^{\mathcal{K}\times T_f \times 2}$.

\noindent
\textbf{Feature Extraction} Using the anchor trajectory database $\mathbf{Y^*}$, we determine the target agent's motion profile by selecting the $K$ most likely future movements as prototypes using historical context as query. This context includes the agent's past ego status, positions of neighboring agents, and environmental layout. We then project the historical embedding into a future trajectories feature space to choose prototypes whose embeddings are close to this projection.

To obtain the embedding of the historical context $Z^{(x)}_i$ for agent $i$, we first convert their historical ego status $X^{(ego)}_i$—the temporal sequence of their velocity and heading in polar coordinates—into Fourier features $Z^{(ego)}_i\in\mathbb{R}^{T_o\times D}$ \cite{tancik2020fourier}, where $D$ is the dimension of embedding space. We then compute the relative spatial-temporal positional embedding $R^{(social)}_{i,j}$ first proposed in the QCNet~\cite{zhou2023query} to encode the historical social context using the position of neighboring agents. $Z^{(ego)}_i$, along with $R^{(social)}_{i,j}$, then is injected into an encoder composed of the factorized attention modules across the temporal and social dimensions, where the ego status embeddings are updated with historical and social contexts.
\begin{equation}
Z^{(ego)}_i = F_{\theta}(Z^{(ego)}_i, \{R^{(social)}_{i,j}\}_{\forall j\in N_i}),
\end{equation} where $N_i$ is the neighbor of agent $i$.
We then takes the embedding corresponding to the last observation step $Z^{(ego)}_i(t)\in\mathbb{R}^D$ to represent an agent's history of past movement and social interactions. 

The environmental layout, alongside historical and social contexts, significantly influences pedestrian decision-making. We propose using a distance array $M_i\in\mathbb{R}^{360}$, representing distances to nearby obstacles, to model an agent's local environment. This distance array is encoded into Fourier features and combined with historical data through cross-attention to derive the historical context embedding
\begin{equation}
    Z^{(x)}_i = G_{\theta}(Z^{(ego)}_i(t),M_i).
\end{equation}

To construct the feature space for future trajectory, we normalize and rotate anchor with respect to the last observed position and heading for each agent. We then obtain the anchor trajectory embedding $Z^{(y)}_\kappa=F_{\phi}(Y_\kappa^*)$, where $Y_\kappa^*\in \mathbf{Y^*}$ for all $\kappa\in \mathcal{K}$ anchor trajectories using an LSTM encoder $F_{\phi}$. Both embeddings $Z^{(x)}_i$ and $Z^{(y)}_\kappa$ will capture the motion pattern in their corresponding time horizon. To maintain consistency between these two separate feature spaces (history and future), we further transform the observation embedding $Z^{(x)}_i$ by projecting it into the future trajectory embedding space via an MLP layer $h$ and obtain an updated feature $Z^{(x)'}_i = h(Z^{(x)}_i)$.

\noindent
\textbf{Prototype Selection}
 We compute the matching score using the cosine similarity values between the updated observation feature and the anchor trajectory embeddings $Z_\kappa^{(y)}$. Specifically, we compute
\begin{equation}
S_i = softmax(\{s^{i}_\kappa|\kappa=1,\cdots,\mathcal{K}\}), \ s^i_\kappa= Z^{(x)'\top}_iZ_\kappa^{(y)},
\label{eq: score}
\end{equation}
where $S_i$ is the set of matching scores between agent $i$'s observed trajectory against all anchor trajectories. To prevent the selection of anchor trajectories that violate the environmental constraint, we perform zero-out operations to those degenerate ones.\footnote{ Since anchor trajectories are fixed, we can pre-label the ones that violate environmental constraint for target agent $i$.} We then pick the top-$K$ highest scored prototype as the prototype trajectory set for agent $i$. 

 We enhance prediction accuracy by refining prototypes with agent's historical embeddings. For agent $i$, the customized prototype is derived by adding its observation embedding to the $k$-th likely prototype $Z_{i_k}^{(y)}$, then decoding it with an LSTM network:
\begin{equation}
    \tilde{Y}_{i_k}^* = LSTM(Z^{(x)}_i + Z_{i_k}^{(y)}).
\end{equation}
This method constructs a customized prototype trajectory set $\mathbf{\tilde{Y}}_i^{*}$ for agent $i$, which serves as a candidate set for joint alignment.

\subsection{Joint Scene Prediction Alignment}
Despite the usage of multi-agent contexts, the prototype selection process still lacks consideration and modeling of the future interaction between agent. Using the highest scoring prototype trajectories of each agent, we can construct an interaction graph $\mathcal{G}$ that estimates the relationship between agents in the future horizon. We then infer the joint distribution of future trajectories $P(Y|X,M)$ over the selected prototypes for all agents in the scene by constructing a Markov Random Field (MRF) from the interaction graph. Inspired by the energy-based formulation of JFP~\cite{Luo2022-qy}, we define the joint trajectory distribution 
\begin{equation}
    P(Y|X,M) = \frac{1}{\mathcal{Z}}\exp(E(Y|X,M)),
\end{equation}
where $\mathcal{Z}$ is the normalizing constant. The energy function is defined to be the sum of unary and pair-wise potentials as:
\begin{multline}
    E(Y|X,E) = \sum_i E_{unary}(Y_i|X,E) \\ 
    + \sum_{(i,j)\in \mathcal{G}}E_{pairwise}(Y_i,Y_j).
\end{multline}

For the unary potential $E_i = E_{unary}(Y_i|X,E)$, we directly use the logit of the prototype selection scores $s_k^i$ indicated in \Cref{eq: score}, where the higher value means that the agent has a higher probability of taking such action. We train a dedicated module to estimate the pairwise potential $E_{ij} =E_{pairwise}(Y_i,Y_j) \in \mathbb{R}^{K\times K}$ for each connected edge between two interacting agents $i$ and $j$, where
\begin{equation}
\label{eq:eq_pair}
    E_{ij} = MLP(\tilde{y}_{i}^*,\tilde{y}_{j@i}^*),
\end{equation}
and $\tilde{y}_{i}^*,\tilde{y}_{j@i}^*$ are the embedding of the selected prototype of $i$ and $j$ in the coordinate of the agent $i$'. We then mask-out the pair-wise potential value of the colliding pairs of prototypes between agent $i$ and $j$ by assigning large negative values. 

\noindent
\textbf{Joint Alignment via Gibb's Sampling} We apply Gibb's sampling to jointly sample and align prototypes selected from the previous step from the joint distribution estimated by the MRF module. We initialize the sampling process using initial samples from the marginal distribution characterized by the unary potential as:
\begin{equation}
    Y^{(0)} \sim P_{unary},\quad P_{unary}=\frac{1}{\mathcal{Z}}\exp(E_{unary}(Y|X,M)).
\end{equation}

 We then iteratively sample for each agent $i$. At each sampling step $\tau$, we sample the trajectory of the agent $i$ by conditioning the sample of the other agents' from the previous step as:
\begin{equation}
    Y_i^{(\tau)}\sim P\left( Y_i | Y^{(\tau-1)}_j, j\neq i \right).
\end{equation}
To sample $K$ sets of scene prediction, we save the samples after $B$ steps of the burn-in period. We present the algorithm for the full sampling process in the supplementary material.

\subsection{Training Objectives}
We train the full pipeline end-to-end using separate loss components for each module. For anchor selection, we apply a focal loss by identifying the prototype with the smallest displacement from the ground-truth trajectory as the GT prototype. The focal loss for the GT prototype's matching score $s_m^i$ is given by:
\[
\mathcal{L}_{focal} = -\alpha(1-s_m^i)^{\gamma}\log(s_m^i),
\]
where $\alpha$ is the balancing parameter and $\gamma$ controls level of loss contribution from the easier samples.

We train the prototype refinement module using a winner-takes-all strategy. Given the refined prototype trajectories $ \tilde{Y}_{i_k}^*$ for $k=1,\cdots,K$, we define the regression loss that penalizes the displacement error of the most accurate prototype,
\begin{equation}
    \mathcal{L}_{regress} = \min_{k=1}^K \|\tilde{Y}_{i_k}^*-Y_i \|_2,
\end{equation}
where $Y_i$ is the GT future trajectory for agent $i$.

To learn a meaningful feature space for future trajectories, we train the future encoder and decoder by reconstructing GT trajectory $Y_i$ via a reconstruction loss
\begin{equation}
    \mathcal{L}_{recon} = \| LSTM(Z^{(x)}_i+F_{\phi}(Y_i))-Y_i \|_2.
\end{equation}
Finally, we make sure that the GT prototype trajectory embedding $Z_m^i$ aligns well with the GT trajectory embedding as
\begin{equation}
    \mathcal{L}_{embed} = \| F_{\phi}(Y_i)-Z_m^i \|_2.
\end{equation}
Therefore, we define the ACP training loss as a weighted sum, 
\begin{equation}
    \mathcal{L}_{ACP} = \beta_1 \mathcal{L}_{focal}+\mathcal{L}_{pred}+\mathcal{L}_{recon}+\mathcal{L}_{embed},
\end{equation}
where $\beta_{1}$ is a hyperparameter to scale the focal loss. We use $\beta_{1}=100$ for the experiments.

To train the pairwise potential, for each edge in the MRF we mark the prototype pairs between agent $i$ and $j$ with the minimum joint displacement error from the ground-truth as the GT label and then compute the focal loss on the pairwise potential value,
\begin{equation}
    \mathcal{L}_{pairwise} = -\alpha(1-e_{ij}^{(m,n)})^{\lambda}log(e_{ij}^{(m,n)}),
\end{equation}
here we assume agent $i$'s $m$th prototype and agent $j$'s $n$th prototype shall produce the minimum joint displacement from their GT future.
Our final loss function is then composed of both the ACP and pairwise potential loss, $\mathcal{L}=\mathcal{L}_{ACP}+\mathcal{L}_{pairwise}$. Note that, for all focal loss, we use $\alpha=0.25$ and $\lambda=2$. We also stop the gradient for $\tilde{y}_{i}^*,\tilde{y}_{j@i}^*$ in \Cref{eq:eq_pair} for the pairwise focal loss.

\section{Experiments}
\label{sec:experiment}

\subsection{Quantitative Evaluation}
\label{sec:quantitative}


\textbf{Dataset}
We evaluated the performance of our suggested prediction algorithm by performing a benchmark using the widely recognized ETH-UCY datasets \cite{lerner2007crowds, pellegrini2009you}. The ETH-UCY dataset contains 1,536 pedestrians from five different scenes: ETH, Hotel, Univ, Zara1, and Zara2; the trajectories are annotated from top-down view footage of surveillance cameras in world coordinates. Here, all results are measured in units of meters. We follow the common training and testing convention first used in Social-GAN \cite{gupta2018social}, with an observation window of 3.2 seconds (8 frames) and a prediction horizon of 4.8 seconds (12 frames) and performed cross-validation against each scene to split the whole dataset into training and testing subsets.

\noindent
\textbf{Evaluation Protocols}
We perform comprehensive evaluations for all benchmarked models based on two main criteria: accuracy and feasibility. 
To capture the stochastic nature of human movement, we sample $k$ samples from each model and perform the evaluation with $k=20$.  We used Joint ADE/FDE (JADE/JFDE)~\cite{weng2023joint} to measure the model's ability to perform joint prediction across all scene agents. 
We measure the feasibility of the predicted trajectories by looking at the environmental and agent-to-agent (A2A) collision rate. We also include the commonly used $min_k$ADE/FDE to evaluate the upper bound of each model's predictive capability by measuring the displacement error of the most accurate sample. We provide detailed definition of these metric in the supplementary material.

\begin{table}[t]
\centering
\caption{$\text{min}_k\text{JADE/JFDE}$ on ETH-UCY dataset with $k=20$. The
best performance is boldfaced and the 2nd place is marked as blue.}
\resizebox{\columnwidth}{!}{
\begin{tabular}{c|ccccc}
\toprule
Model & ETH & Hotel & UNIV & ZARA1 & ZARA 2 \\
Avg. \# Agents&	2.6&	3.5&	25.7&	3.7&	6.3\\
\midrule
Agentformer&	0.619/1.136&	0.303/0.603&	0.622/1.311&	\textbf{0.325/0.660}&	0.314/0.663 \\
GP-Graph&	0.588/1.003&	0.307/0.604&	0.623/1.319&	\textcolor{blue}{0.342/0.709}&	0.322/0.690\\
EqMotion&	0.547/0.822&	\textbf{0.230/0.405}&	\textbf{0.499/1.087}&	0.367/0.775&	\textcolor{blue}{0.299/0.692}\\
SingularTrajectory&	\textcolor{blue}{0.462/0.714}&	0.286/0.553&	0.668/1.395&	0.376/0.747&	0.356/0.747\\
\midrule
Ours & 0.704/1.226 & \textcolor{blue}{0.229/0.420} & 0.715/1.472 & 0.361/0.724 & \textbf{0.304/0.623} \\
\bottomrule
\end{tabular}
}
\label{tab:jade}
\end{table}

\begin{table}[t]
\centering
\caption{Agent-to-Agent Collision Rate ($\text{threhold}=0.2$m) on ETH-UCY dataset. The
best performance is boldfaced and the 2nd place is marked as blue.}
\resizebox{\columnwidth}{!}{
\begin{tabular}{c|ccccc}
\toprule
Model & ETH & Hotel & UNIV & ZARA1 & ZARA 2 \\
Avg. \# Agents&	2.6&	3.5&	25.7&	3.7&	6.3\\
\midrule
Ground Truth&	0&	0.001&	0.035&	0&	0.007\\
\midrule
Agentformer&	0.056&	\textcolor{blue}{0.031}&	0.205&	\textcolor{blue}{0.024}&	\textcolor{blue}{0.064}\\
GP-Graph&	\textcolor{blue}{0.033}&	\textcolor{blue}{0.031}&	\textcolor{blue}{0.179}&	0.031&	0.065\\
EqMotion&	0.52&	0.047&	0.230&	0.144&	0.108\\
SingularTrajectory&	0.066&	0.077&	0.217&	0.064&	0.133\\
\midrule
Ours & \textbf{0.00}& \textbf{0.001} & \textbf{0.006}& \textbf{0.001} & \textbf{0.001}\\
\bottomrule
\end{tabular}
}
\label{tab:collision}
\end{table}

\begin{table}[t]
\centering
\caption{Environmental Collision Rate on ETH-UCY dataset. The
best performance is boldfaced and the 2nd place is marked as blue.}
\resizebox{\columnwidth}{!}{
\begin{tabular}{c|ccccc}
\toprule
Model & ETH & Hotel & UNIV & ZARA1 & ZARA 2 \\
\midrule
Ground Truth&	0&	0&	0.022&	0&	0.002 \\
\midrule
Agentformer&	0.32&	0.147&	0.057&	0.049& \textcolor{blue}{0.039}\\
GP-Graph&	0.258&	0.071&	0.09&	0.051&	0.056\\
EqMotion&	0.677&	0.093&	0.126&	0.076&	0.067\\
SingularTrajectory&	\textcolor{blue}{0.084}&	\textcolor{blue}{0.074}&	\textcolor{blue}{0.067}&	\textcolor{blue}{0.033}&	0.038\\
\midrule
Ours & \textbf{0.062}& \textbf{0.031}& \textbf{0.009}& \textbf{0.024} & \textbf{0.009} \\
\bottomrule
\end{tabular}
}
\label{tab:env_violation}
\end{table}

\begin{table}[t]
\centering
\caption{$\text{min}_k\text{ADE/FDE}$ on ETH-UCY dataset with $k=20$. The
best performance is boldfaced and the 2nd place is marked as blue.}
\resizebox{\columnwidth}{!}{
\begin{tabular}{c|ccccc}
\toprule
Model & ETH & Hotel & UNIV & ZARA1 & ZARA 2 \\
\midrule
Agentformer&	0.45/0.79&	0.14/0.22&	0.25/0.45&	\textcolor{blue}{0.18/0.30}&	\textcolor{blue}{0.14/0.23} \\
GP-Graph&	0.43/0.64&	0.18/0.30&	\textcolor{blue}{0.24/0.42}&	\textbf{0.17/0.31}&	0.15/0.29\\
EqMotion&	0.50/0.72&	\textcolor{blue}{0.13/0.18}&	\textbf{0.23/0.43}&	0.20/0.37&	\textbf{0.13/0.23}\\
SingularTrajectory&	\textcolor{blue}{0.35/0.42}&	0.13/0.19&	0.25/0.43&	0.18/0.38&	0.14/0.25\\
\midrule
Ours & 0.59/1.06& 0.15/0.25 & 0.41/0.82& 0.20/0.39 & 0.17/0.32 \\
\bottomrule
\end{tabular}
}
\label{tab:ade}
\end{table}


\begin{figure*}[t]
    \begin{subfigure}[b]{0.196\textwidth}
    \centering
        \includegraphics[width=\linewidth]{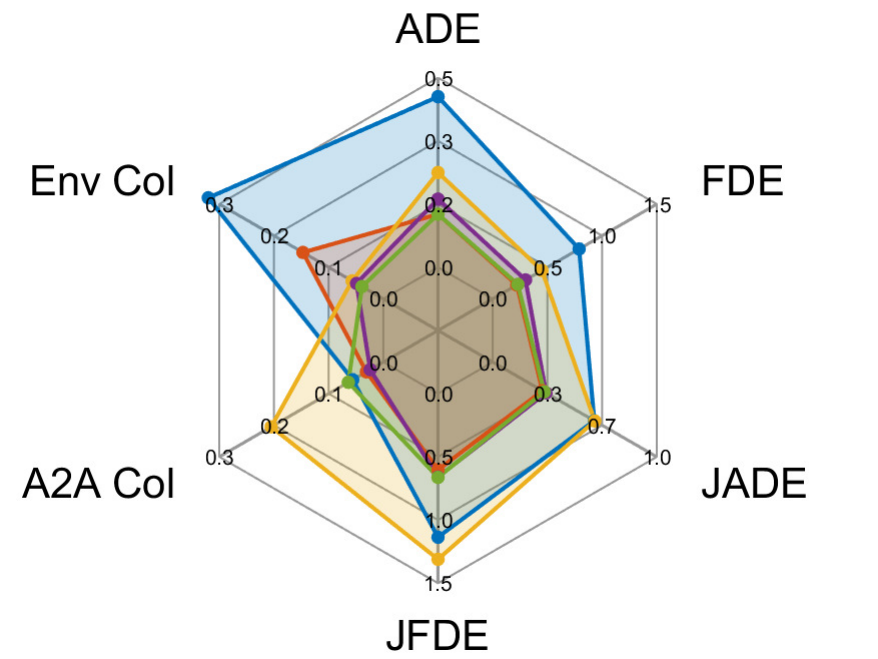}    
        \caption{AgentFormer}
        \end{subfigure}
        \hfill
    \begin{subfigure}[b]{0.196\textwidth}
        \centering
        \includegraphics[width=\linewidth]{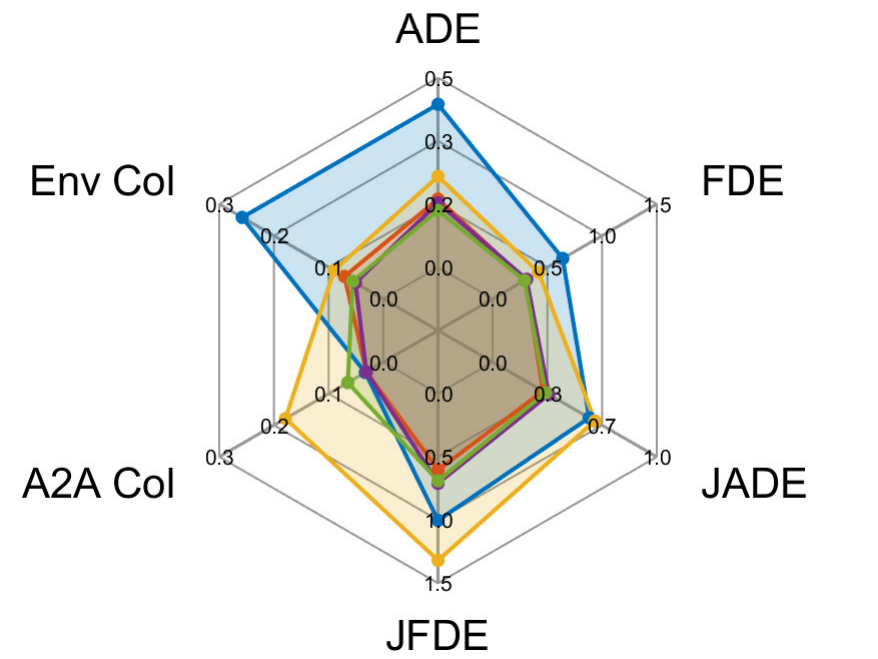}
        \caption{GP-Graph}
        \end{subfigure}
    \hfill
    \begin{subfigure}[b]{0.196\textwidth}
        \centering
        \includegraphics[width=\linewidth]{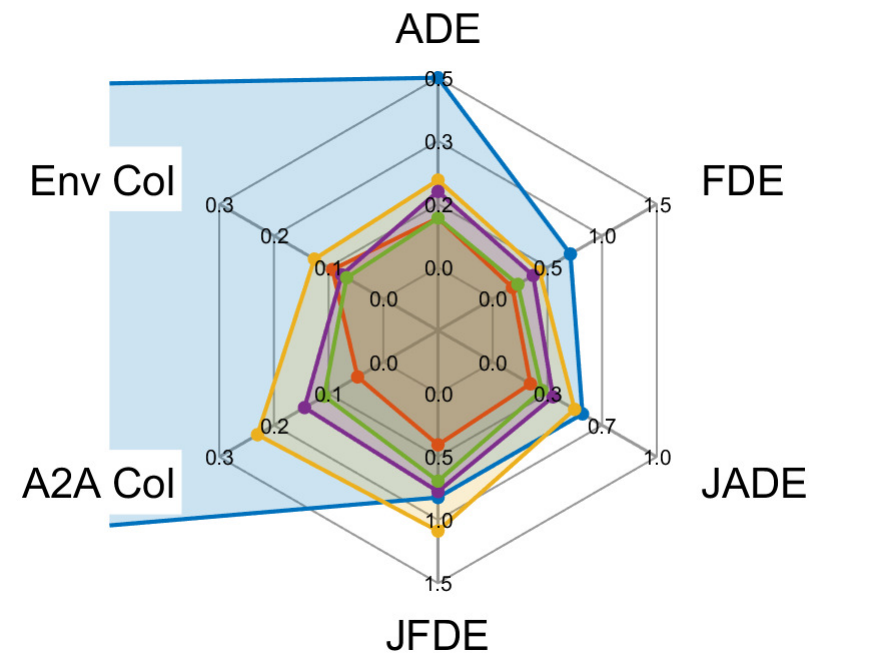}
        \caption{EqMotion}
        \end{subfigure}
    \hfill
    \begin{subfigure}[b]{0.196\textwidth}
        \centering
        \includegraphics[width=\linewidth]{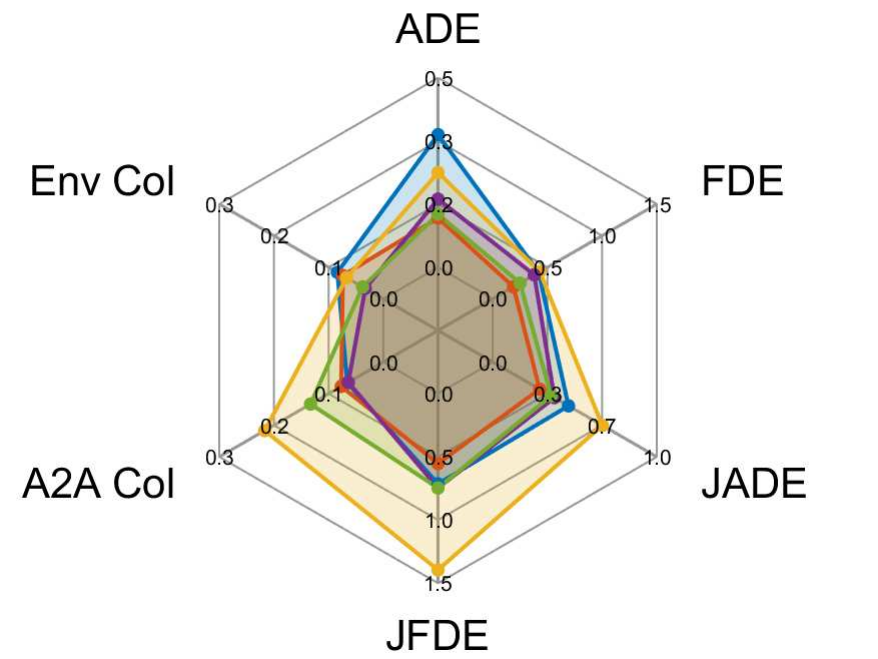}
        \caption{SingularTrajectory}
        \end{subfigure}
    \hfill
    \begin{subfigure}[b]{0.196\textwidth}
        \centering
        \includegraphics[width=\linewidth]{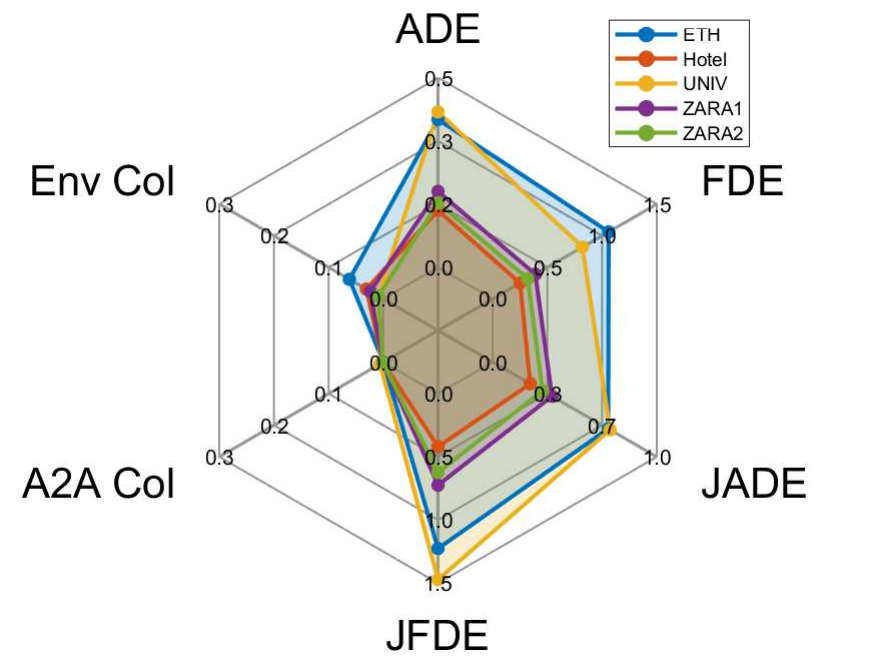}
        \caption{Ours}
        \end{subfigure}    
    \vspace{-0.2in}
    \caption{Radar plot for all evaluation metrics among the five testing splits of ETH-UCY dataset. }
    \label{fig:radars}
    \vspace{-0.2in}
\end{figure*}

\begin{table*}[t]
\centering
\caption{Ablation study on environmental and social collision rate. The best performance is boldfaced and the 2nd place is marked as blue.}
\vspace{-0.1in}
\label{tab:aba_col}
\resizebox{\textwidth}{!}{
\begin{tabular}{ccc|ccccc|ccccc}
\toprule
\multicolumn{3}{c|}{Model} & ETH & HOTEL & UNIV & ZARA1 & ZARA2 & ETH & HOTEL & UNIV & ZARA1 & ZARA2 \\
\multicolumn{3}{c|}{Avg. \# Agents} & 2.6 & 3.5 & 25.7 & 3.7 & 6.3 & 2.6 & 3.5 & 25.7 & 3.7 & 6.3 \\
\midrule
Env Filter & A2A Filter & Gibbs & \multicolumn{5}{c|}{Environmental Collision} & \multicolumn{5}{c}{A2A Collision}  \\
\midrule
$\times$ & $\times$ & $\times$ & 0.154 & 0.116 & 0.039 & 0.077 & 0.053 & \textcolor{blue}{0.038} & \textcolor{blue}{0.054} & \textcolor{blue}{0.250} & \textcolor{blue}{0.074} & 0.130 \\
\midrule
$\checkmark$ & $\times$ & $\times$ & \textcolor{blue}{0.060} & \textcolor{blue}{0.037} & \textcolor{blue}{0.012} & \textcolor{blue}{0.031} & 0.020 & 0.067 & 0.060 & 0.253 & 0.078 & \textcolor{blue}{0.119} \\
$\checkmark$ & $\checkmark$ & $\times$ & \textbf{0.058} & 0.046 & 0.013 & 0.034 & \textcolor{blue}{0.012} & 0.066 & 0.072 & 0.253 & 0.082 & 0.123 \\
$\checkmark$ & $\checkmark$ & $\checkmark$ & 0.062 & \textbf{0.031} & \textbf{0.009} & \textbf{0.024} & \textbf{0.009} & \textbf{0.000} & \textbf{0.001} & \textbf{0.006} & \textbf{0.001} & \textbf{0.001} \\
\bottomrule
\end{tabular}
}
\vspace{-0.2in}
\end{table*}

\begin{table}[t]
\centering
\caption{Ablation study on JADE/JFDE metric. The best performance is boldfaced and the 2nd place is marked as blue.}
\vspace{-0.1in}
\resizebox{\columnwidth}{!}{
\begin{tabular}{ccc|ccccc}
\toprule
\multicolumn{3}{c|}{Model} & ETH & HOTEL & UNIV & ZARA1 & ZARA2 \\
\multicolumn{3}{c|}{Avg. \# Agents} & 2.6 & 3.5 & 25.7 & 3.7 & 6.3 \\
\midrule
Env Filter & A2A Filter & Gibbs &  \\ 
$\times$ & $\times$ & $\times$ & 0.703/1.225 & 0.246/0.446 & 0.702/1.447 & 0.397/0.796 & 0.328/0.672 \\
\midrule
$\checkmark$ & $\times$ & $\times$ & 0.783/1.305 & 0.318/0.548 & 0.626/1.294 & 0.380/0.766 & 0.321/0.670 \\
$\checkmark$ & $\checkmark$ & $\times$ & 0.707/1.234 & 0.256/0.478 & 0.680/1.406 & 0.372/0.744 & 0.308/0.639 \\
$\checkmark$ & $\checkmark$ & $\checkmark$ & 0.704/1.226 & \textbf{0.229/0.420} & 0.715/1.472 & \textbf{0.361/0.724} & \textbf{0.304/0.623} \\
\bottomrule
\end{tabular}
}
\vspace{-0.2in}
\label{tab:aba_jade}
\end{table}

\noindent
\textbf{Analysis}
We evaluate trajectory prediction using the ETH-UCY dataset and compare our method with Agentformer \cite{yuan2021agentformer}, GP-Graph \cite{bae2022learning}, EqMotion \cite{xu2023eqmotion}, and SingularTrajectory \cite{bae2024singulartrajectory}. We assess their JADE/JFDE performance, environmental, and social collision rates by replicating the benchmarked model with the original publication's checkpoints.the 

\noindent
\textbf{Joint Displacement Error} \Cref{tab:jade} presents experiment results of JADE/JFDE on the five partitions of the ETH-UCY dataset. JADE/JFDE evaluates model's ability to simultaneously produce accurate predictions for all scene agents, a more challenging metric compared with the commonly used $min_k$ADE/FDE. The metric requires the model not only to produce accurate individual predictions, but also to align the best samples for all scene agents. Our proposed method demonstrates the ability to match and surpass the overall performance of the SOTA models. Specifically, our model provides a sizable improvement on the Hotel and ZARA2 split. 

\noindent
\textbf{Social Collision Rate} We evaluate the model's understanding of social contexts using the Agent-to-Agent (A2A) Collision Rate, as in \Cref{tab:collision}, applying a $0.2$m collision threshold to match minimal ground truth collisions. ETH, Hotel, and Zara1 contain scenes with sparse social interactions, each with fewer than five agents per scene, while UNIV and ZARA 2 feature denser scenes. Our model achieves almost collision-free trajectory generation across the whole testing splits, outperforming all benchmarked models. Notably, all SOTA models struggle to generate collision-free predictions, particularly in UNIV and ZARA 2 splits, where up to 20\% of the generated trajectories include collisions. This highlights a gap in HTP research, where despite the extensive usage of multi-agent contexts, SOTA models still fail to fully capture socially compliant behaviors. Our method addresses this issue by integrating learned pair-wise potential, collision filtering, and effective sampling strategies to produce trajectories that accurately reflect agents' intentions and maintain social feasibility.

\noindent
\textbf{Environmental Violation} \Cref{tab:env_violation} shows the environmental violation rate in the model predictions for the ETH-UCY dataset. The environmental layout represents a significant decision factor in pedestrian movement. In stochastic trajectory prediction, where the model produces multiple prediction samples, all outputs should follow the environmental layout to be viable for downstream tasks. We used the ETH-UCY map annotated by \cite{mangalam2021goals} as the gold standard. Both SingularTrajectory and our method incorporate environmental information. Our proposed model achieved minimal environmental collisions using filtering techniques during prototype selection phase, which improved SOTA performance. Compared with SingularTrajectory, which performs environmental correction directly on model predictions, our filtering and refinement approach proved more effective, reducing collisions across the dataset.

\Cref{fig:radars} shows a radar plot of all evaluation metrics, highlighting our model's significant improvement in feasibility metrics, especially collision avoidance, which is essential for real-world deployment. This is achieved with a minor trade-off in predictive accuracy, reflected by a slight decrease in standard ADE and FDE scores. While this compromise is also evident in $\min_k\text{ADE}/\text{FDE}$ in \Cref{tab:ade}, we emphasize that $\min_k\text{ADE}/\text{FDE}$ measures only the upper bound of model capability. We strategically prioritize generating a single, highly plausible, and compliant trajectory over optimizing for best-case statistical accuracy, as this better reflects practical application constraints. Further detailed analysis is provided in the supplementary material.


\begin{figure*}[t]
    \centering
    \includegraphics[width=1\linewidth]{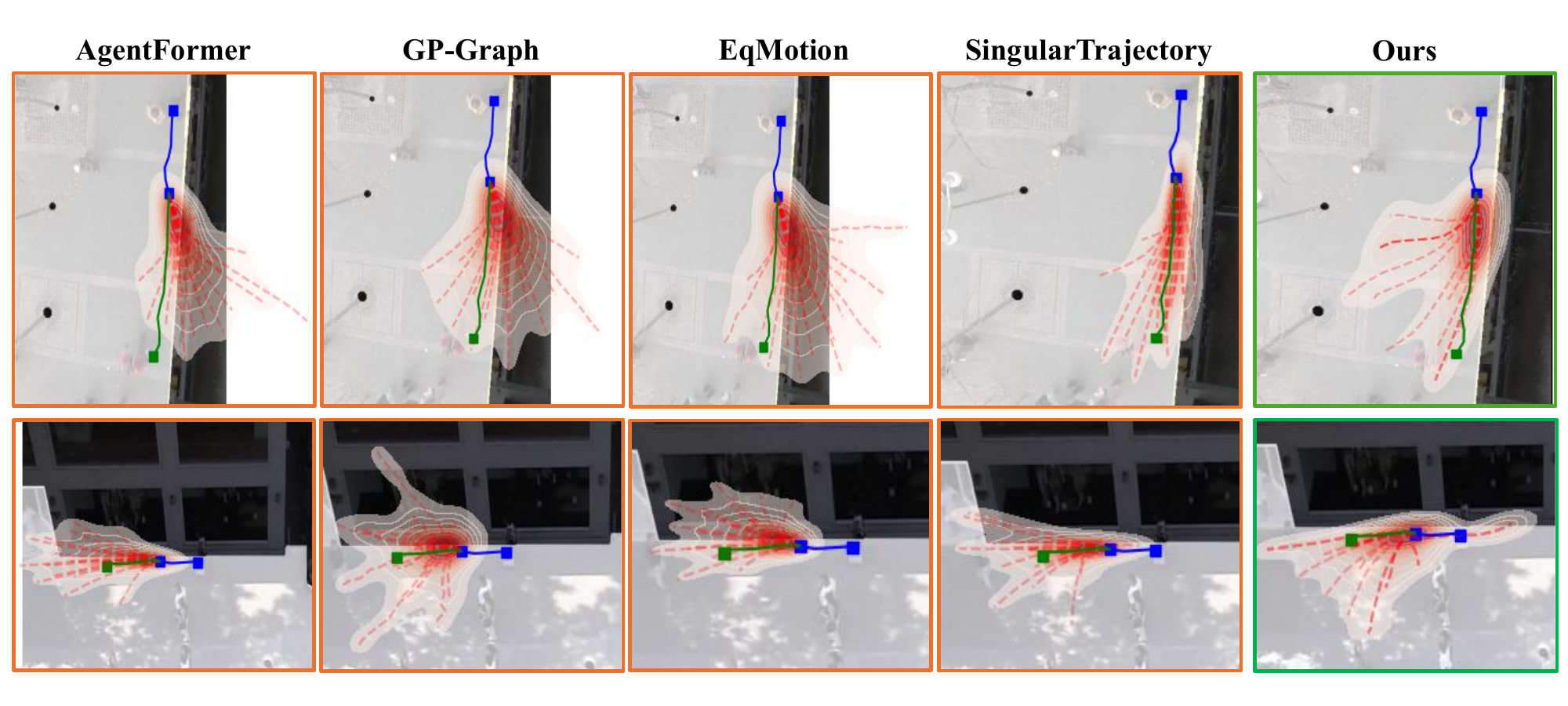}
    \caption{Visualization for all prediction for two individual agents from Hotel (Top) and Zara1 (bottom) splits of ETH-UCY Dataset. \newline \textcolor{Blue}{Blue line:Historical Trajectory}, \textcolor{Green}{Green line: Ground Truth Future Trajectory}, \textcolor{red}{Red dashed line: Predictions}.}
    \label{fig:qual1_env}
    \vspace{-0.1in}
\end{figure*}

\begin{figure*}[t]
    \centering
    \includegraphics[width=1\linewidth]{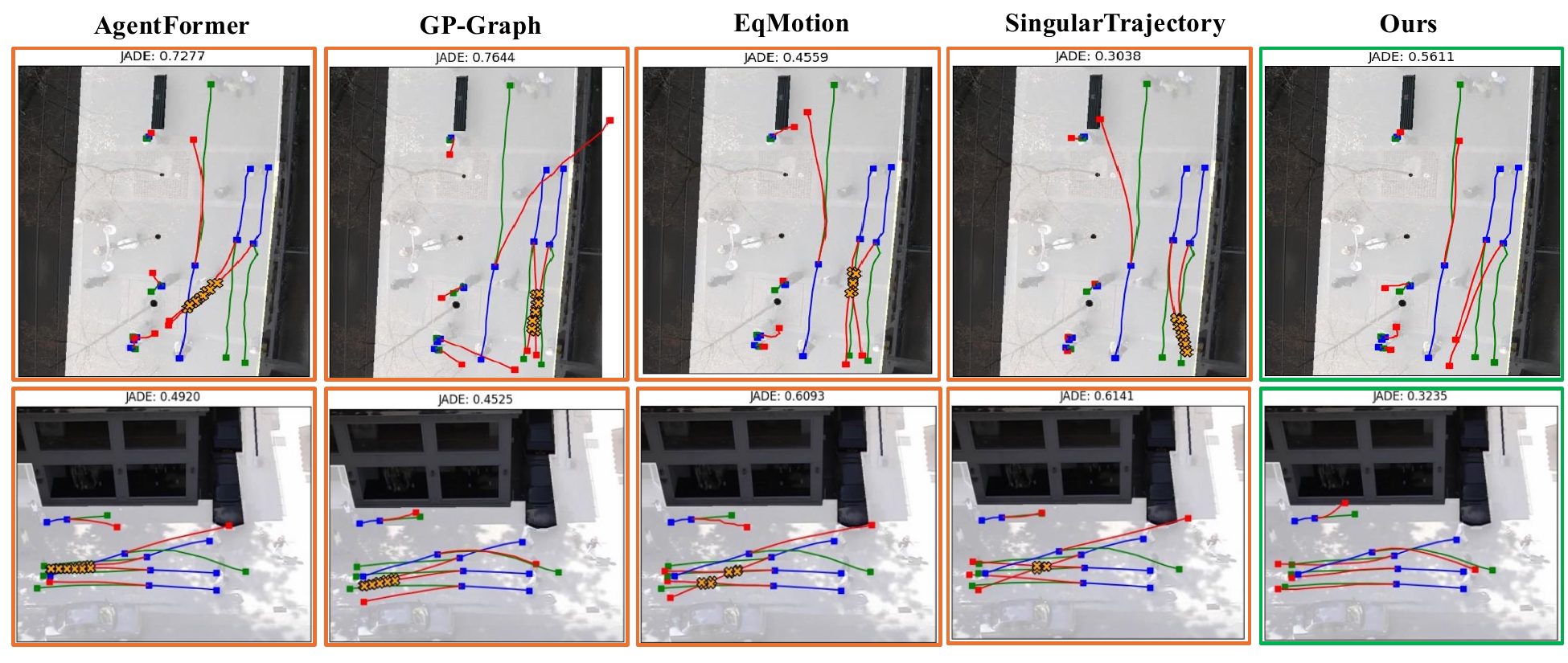}
    \caption{Scene prediction with best JADE performance from Hotel (top) and Zara2 (bottom) split. \textcolor{Blue}{Blue line:Historical Trajectory}, \textcolor{Green}{Green line: Ground Truth Future Trajectory}, \textcolor{red}{Red dashed line: Prediction}, \textcolor{brown}{Yellow Cross: Agent-to-Agent Collision}.}
    \label{fig:qual2_a2a}
    \vspace{-0.1in}
\end{figure*}

\subsection{Qualitative Analysis}
We provide qualitative analysis with visualizations of the individual and scene predictions to better showcase the behavior and characteristics of the models in social and environmental interactions. 

\Cref{fig:qual1_env} illustrates two examples of individual predictions from the Hotel and Zara1 scenes. Our model effectively produces diverse, scene-compliant trajectories. In the first example (\Cref{fig:qual1_env} top), the sample distribution aligns with the environment, predicting straight or rightward paths (in the agent's perspective) as the agent walks alongside a solid barrier on their left. SingularTrajectory uses map gradients for corrections, resulting in compliant but less diverse predictions focusing on forward movement. Agentformer, GP-Graph, and EqMotion do not comply with environmental constraints, mainly predicting leftward turns, which are impractical and unrealistic. In the second example (\Cref{fig:qual1_env} bottom), our model maintains environmental adherence, generating diverse motions, including 'U-turn' behavior. The other four SOTA methods fail to produce compliant predictions, even for the SingularTrajectory which performs environmental correction; and despite their diverse prediction, they generate infeasible trajectories.

\Cref{fig:qual2_a2a} shows two scene predictions from Hotel and Zara1, chosen by the optimal JADE performance among each model's 20 samples. Though SOTA HTP models excel at individual predictions ($min_kADE$ metric), accurately predicting all agents in a scene remains challenging. Due to the harder nature of joint prediction, model evaluation should allow the generation of alternative future as long as the scene predictions are compliant. In the first example (\Cref{fig:qual2_a2a} top), our method mostly aligns with agents' original paths, except for two parallel agents on the right. Our model predicts collision-free paths while keeping their formation, unlike other SOTA models which fail to do so without causing overlaps. In the second example (\Cref{fig:qual2_a2a} bottom), our model excels in social comprehension, accurately predicting the intent of all four agents shown in the center of the image, achieving the lowest JADE among benchmarks models, which either miss intentions or predict collisions.

\subsection{Ablation Study}
\label{subsec:aba}
\Cref{tab:aba_col} and \Cref{tab:aba_jade} show the ablation study on environment and social filtering, along with the Gibb's sampling module, regarding collision and joint prediction accuracy metrics. Environmental filtering notably reduces collisions, enhancing model compliance. Without this filtering, the model struggles to maintain environmentally compliant predictions despite using environmental context for prototype refinement.

A2A filtering alone in pair-wise potential is insufficient for socially compliant predictions. We use belief propagation on the MRF to update the unary potential for each agent. We then re-rank the chosen prototype by the updated unary potential values and align the scene prediction using these ranks, (e.g., align all rank 1,2,3...,20 samples). We observe that re-ranking alone does not prevent social collisions or enhance JADE/JFDE performance. Combining A2A filtering with Gibbs sampling significantly reduces A2A collisions and improves joint prediction accuracy, showcasing the importance of sampling from a joint trajectory distribution.

\section{Conclusion}
\label{sec:conclusion}
In this paper, we present JACoP, a novel multi-stage framework that enhances stochastic Human Trajectory Prediction (HTP) by focusing on collective scene-level plausibility. Unlike current top models that often overlook scene consistency, JACoP ensures joint trajectory prediction with minimal breaches of environmental and social constraints. Evaluations show JACoP performs excellently in collective compliance, significantly reducing collisions and boundary violations while maintaining high predictive accuracy. Despite minor declines in accuracy in some data sets and higher time complexity during sampling, JACoP's near-perfect scene compliance makes it suitable for tasks like simulation or crowd behavior analysis, which have less stringent real-time needs.

\newpage

{
    \small
    \bibliographystyle{ieeenat_fullname}
    \bibliography{main}
}

\clearpage
\setcounter{page}{1}
\setcounter{section}{0}
\renewcommand*{\thesection}{\Alph{section}}
\renewcommand*{\thepage}{S\arabic{page}}
\providecommand*{\theHpage}{}
\renewcommand*{\theHpage}{supplement.\arabic{page}}
\renewcommand*{\theHsection}{supplement.\Alph{section}}
\renewcommand*{\theHfigure}{supplement.\arabic{figure}}
\renewcommand*{\theHtable}{supplement.\arabic{table}}
\renewcommand*{\theHequation}{supplement.\arabic{equation}}
\renewcommand*{\theHalgorithm}{supplement.\arabic{algorithm}}
\maketitlesupplementary

\begin{table*}[t]
\centering
\caption{Average ADE/FDE and KDE-based NLL Benchmark. The best performance is boldfaced and the 2nd place is marked as blue.}
\vspace{-0.1in}
\label{tab:supp_avgade_nll}
\resizebox{\textwidth}{!}{
\begin{tabular}{c|ccccc|ccccc}
\toprule
Model & ETH & HOTEL & UNIV & ZARA1 & ZARA2 & ETH & HOTEL & UNIV & ZARA1 & ZARA2 \\
Avg. \# Agents & 2.6 & 3.5 & 25.7 & 3.7 & 6.3 & 2.6 & 3.5 & 25.7 & 3.7 & 6.3 \\
\midrule
Model & \multicolumn{5}{c|}{Average ADE/FDE} & \multicolumn{5}{c}{KDE NLL}  \\
\midrule
AgentFormer & 1.99/4.39	& 0.87/2.01	& 1.01/2.26	& 0.72/1.63  & 0.74/1.71 & 2.70	& 1.60	&	1.88&  \textcolor{blue}{1.52} & 1.53\\
GP-Graph& \textcolor{blue}{1.21/2.51}	& 0.71/1.53	& \textbf{0.91/1.31}	& \textcolor{blue}{0.68/1.50}  & \textcolor{blue}{0.57/1.27} & 2.31	& 1.51	& \textcolor{blue}{1.80}	& 1.52  &\textcolor{blue}{1.42}	\\
EqMotion& 1.42/2.99 & \textcolor{blue}{0.64/1.32}	& 2.30/5.39	& 0.82/1.83	& 0.68/1.54  & \textcolor{blue}{2.23} & \textcolor{blue}{1.37}	& \textbf{1.78}	& 1.60	& 1.47  \\
SingularTrajectory& 1.47/2.77 & 0.89/1.88 & 1.12/2.36 & 0.92/1.96	& 1.02/2.23  &\textbf{2.04} &1.58 & 1.93	&1.67	&1.77  \\
JACoP &	\textbf{1.03/1.97}& \textbf{0.50/1.05} & \textcolor{blue}{0.82/1.70} & \textbf{0.59/1.26}  & \textbf{0.46/0.99} &	2.54&\textbf{1.28}& 2.07	&\textbf{1.44}  &\textbf{1.35}\\
\bottomrule
\end{tabular}
}
\end{table*}

\Cref{sec:supp-sampling} details our sampling algorithm for generating scene predictions via prototype alignment. \Cref{sec:supp-eval-metric} defines the evaluation metrics used in our benchmark experiments. \Cref{sec:supp-avg-ade} provide an additional benchmark on the average ADE/FDE and KDE NLL metrics, highlighting JACoP's ability to produce predictions that are tightly concentrated around the ground-truth future and better align with agent intentions. \Cref{sec:supp-alt-baseline} experiment on alternative agent-centric profiler baseline, confirming our proposed ACP's superiority in accuracy and collision metrics. \Cref{sec:supp-add-qual} provides more qualitative analysis reflecting JACoP's ability to produce accurate, collision-free predictions in complex scenes. \Cref{sec:supp-sdd} provide additional benchmark on SDD dataset. \Cref{sec:supp-col-vs-num-agent} provides analysis on collision rate and number of agent per scene.

\section{Sampling Algorithm}
\label{sec:supp-sampling}
\Cref{alg:gibbs} details our proposed sampling algorithm, which generates the scene prediction by aligning the prototype trajectories selected by our Agent-Centric Profiler (ACP). We employ Gibbs sampling, iteratively sampling a trajectory for each agent conditioned on the trajectories of all other agents from the previous sampling step. The process is initialized by sampling from the marginal distribution inferred for each agent by the ACP module. We perform a total of $B+K$ iterations across all agents and retain the samples from the final $K$ steps as our set of scene predictions. This sampling procedure ensures that the samples converge to the distribution defined by the Markov Random Field (MRF) and the social filtering process, which is essential for achieving realistic predictions with no social collisions and high joint accuracy.

The sampling algorithm requires a time complexity of $O(N(B+K))$, where the overall complexity grows linearly with the number of agents ($N$). We can further parallelize the sampling process by initializing $K$ independent processes, which allows the time complexity to be reduced to $O(BN)$.

\section{Evaluation Metrics}
\label{sec:supp-eval-metric}
\paragraph{Minimum ADE and FDE} The commonly used metric $min_kADE/FDE$ evaluates the average and final displacement error of the most accurate sample among the multi-modal prediction outputs. The ground truth future is denoted as $Y=\{Y_1,\cdots,Y_n\}$, representing a scene comprising multiple agents. Each agent's trajectory is further detailed as $Y_i=\{y_{i,1},\cdots,y_{i,T_f}\}$. The $K$ prediction samples are represented as $\hat{Y}^{(k)}=\{\hat{Y}^{(k)}_1,\cdots,\hat{Y}^{(k)}_n\}$, with each agent's sample expressed as $\hat{Y}^{(k)}_i=\{\hat{y}^{(k)}_{i,1},\cdots,\hat{y}^{(k)}_{i,T_f}\}$, for each $k=1,...,K$. The metrics are defined as
\begin{multline}
    min_kADE(Y,\hat{Y}) = \frac{1}{TN}\sum_{i=1}^N\min_{k=1}^K\sum_{t=1}^T||y_{t,i}-\hat{y}^{(k)}_{t,i}||_2^2, \\
    min_kFDE(Y,\hat{Y}) = \frac{1}{N}\sum_{i=1}^N\min_{k=1}^K||y_{T_f,i}-\hat{y}^{(k)}_{T_f,i}||_2^2.
\end{multline}

\paragraph{JADE/JFDE} The Joint Accuracy metrics, referred to as JADE/JFDE, were initially introduced in \cite{weng2023joint} with the objective of enhancing the widely used marginal $min_kADE/FDE$ metrics. These joint metrics are designed to more accurately reflect a model's capability to predict the collective future trajectories of all agents present within a given scene. Unlike the conventional approach, which involves selecting each agent's most accurate prediction samples independently, JADE/JFDE calculates the average displacement error across all agents within a single prediction sample. The JADE and JFDE are defined as 
\begin{multline}
    JADE(Y,\hat{Y}) = \frac{1}{TN}\min_{k=1}^K\sum_{i=1}^N\sum_{t=1}^T||y_{t,i}-\hat{y}^{(k)}_{t,i}||_2^2, \\
    JFDE(Y,\hat{Y}) = \frac{1}{N}\min_{k=1}^K\sum_{i=1}^N||y_{T_f,i}-\hat{y}^{(k)}_{T_f,i}||_2^2. 
\end{multline} 

\paragraph{Agent-to-Agent Collision} The agent-to-agent collision rate measures the proportion of predicted trajectories that intersect with another agent's path in the same scene prediction. Two agents are considered to collide if their positions come within $r=0.2$ meters at any future time step. We first define an indicator function for collision as
\begin{multline}
    \mathbb{1}_{col}(\hat{Y}^{(k)}_i,\hat{Y}^{(k)}_j) = 1, \\
    \text{if } \exists \ t, \text{such that } ||\hat{y}^{(k)}_{t,i}-\hat{y}^{(k)}_{t,j}||^2_2<r.
\end{multline}
Then we compute the agent-to-agent collision rate as
\begin{multline}
    \text{A2A}_{CR} = \frac{1}{NK}\sum_{i=1}^N\sum_{k=1}^K \mathbb{1}_{col}(\hat{Y}^{(k)}_i,\hat{Y}^{(k)}_j), \\\text{for any } j, i\neq j.
\end{multline}

\paragraph{Environmental Collision Rate} The environmental collision rate evaluates the percentage of predicted trajectories overlap with non-navigable areas or environmental obstacles. We first define an indicator function for environmental violations given the prediction $\hat{Y}^{(k)}_i$ and navigability map $M$,
\begin{multline}
    \mathbb{1}_{env}(\hat{Y}^{(k)}_i,M) = 1, 
    \text{if } \exists t \text{ such that } M(\hat{y}_{t,i})\neq 1.
\end{multline}
Then we compute the proportion of predictions with environmental violation via
\begin{equation}
    \text{ENV}_{CR}=\frac{1}{NK}\sum_{i=1}^N\sum_{k=1}^K \mathbb{1}_{env}(\hat{Y}^{(k)}_i,M).
\end{equation}

\paragraph{Average ADE/FDE}
Both the minimum $k$ Average/Final Displacement Error ($min_kADE/FDE$) and the Joint Average/Final Displacement Error (JADE/JFDE) primarily evaluate prediction accuracy based on the best marginal or joint sample. However, in a real-world deployment setting, distinguishing the most accurate output from a set of model predictions is impossible. Therefore, it is crucial that the entire set of predicted futures does not deviate excessively from the true trajectory. To better evaluate the model's overall capability in producing realistic future predictions, we propose the Average ADE/FDE as an additional accuracy measure.
\begin{multline}
    avgADE(Y,\hat{Y}) = \frac{1}{TNK}\sum_{i=1}^N\sum_{k=1}^K\sum_{t=1}^T||y_{t,i}-\hat{y}^{(k)}_{t,i}||_2^2, \\
    avgFDE(Y,\hat{Y}) = \frac{1}{NK}\sum_{i=1}^N\sum_{k=1}^K||y_{T_f,i}-\hat{y}^{(k)}_{T_f,i}||_2^2.
\end{multline} 
The key difference between $min_kADE/FDE$ and $avgADE/FDE$ lies in their aggregation strategy: instead of selecting the best sample (i.e., the minimum error) from the $K$ outputs, $avgADE/FDE$ averages the displacement error across all $K$ model outputs. A smaller value for these average metrics indicates that the model's entire distribution of predictions is tightly concentrated around the ground-truth future.

\paragraph{KDE-based NLL}
To further examine the likelihood of each method producing the ground-truth (GT) future among all their prediction outputs, we benchmark the models using the Kernel Density Estimation (KDE)-based Negative Log-Likelihood (NLL) metric, as proposed in \cite{ivanovic2019trajectron, thiede2019analyzing}. The KDE-NLL metric quantifies the probability density of the GT future trajectory under a distribution inferred from the $K$ predicted samples produced by the models. A smaller NLL value indicates a tighter concentration of the model's predicted distribution around the GT future, reflecting a more accurate and confident overall prediction set. Here we use Silverman's rule of thumb to determine the optimal bandwidth for the KDE estimation.

\section{Average ADE/FDE and KDE NLL Analysis}
\label{sec:supp-avg-ade}
Previous accuracy metrics primarily emphasize the model's best sample; this design was originally introduced to compensate for the inherent stochastic nature of human trajectory. While this metric effectively reflects the upper bound of a model's predictive capability, its optimization can inadvertently lead to models learning shortcuts. Specifically, a model may produce highly diverse or scattered predictions hoping that at least one sample happens to align closely with the ground-truth future. Our previous benchmark results have shown that state-of-the-art (SOTA) models, despite their good accuracy upper bound, frequently produce highly unlikely predictions that violate social and environmental constraints. Consequently, these best-case accuracy measures become less convincing as a true indicator of model quality.

To address this limitation, we introduce two additional benchmark experiments to provide a more complete assessment of the overall model capability for producing accurate predictions. Specifically, we measure the Average ADE/FDE and KDE-NLL of the model outputs, as presented in \Cref{tab:supp_avgade_nll}. Our JACoP model achieves a notable improvement in Average ADE/FDE performance compared with state-of-the-art (SOTA) methods. This suggests that the model's output distribution, on average, more closely aligns with the ground-truth (GT) agent intentions.

To further support this finding, we evaluate the KDE-NLL metric, which measures the likelihood of the GT future trajectory under the sampling distribution of the model outputs. Our model demonstrates strong performance relative to SOTA methods, particularly on the Hotel, Zara1, and Zara2 splits, where JACoP sets a new performance benchmark.

Taken together, these results—combined with JACoP's ability to generate trajectories with minimal scene violations—support the conclusion that our JACoP model functions as a reliable trajectory predictor capable of generating multiple future possibilities that better reconcile both GT agent intentions and environmental and social contexts.

\begin{figure}[t]
\includegraphics[width=\linewidth]{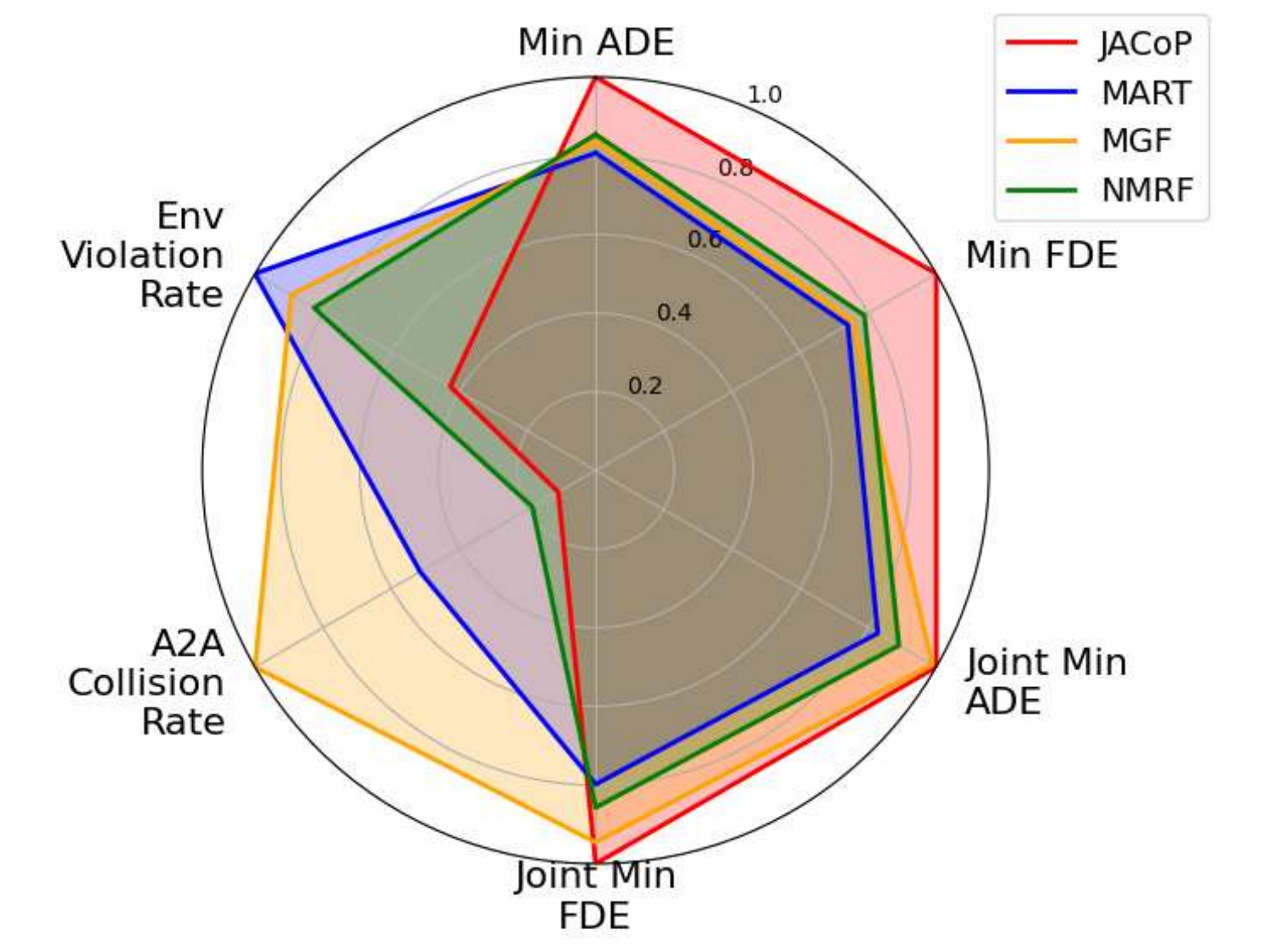}
\vspace{-0.25in}
\caption{Radar Chart for normalized SDD evaluation result}
\label{fig:sdd-radar}
\end{figure}

\begin{table}[t]
\centering
\caption{Accuracy and Collision Benchmark on SDD dataset. $min_kADE$ and JADE measured in pixel space. A2A collision threshold equals 5 pixels (roughly 0.2m). Best result in \textbf{bold face} and second best marked by \textcolor{blue}{blue} color}
\resizebox{\columnwidth}{!}{
\begin{tabular}{c|cccc}
\toprule
Model & $min_kADE/min_kADE$ & JADE/JFDE & A2A Collision & A2E Collision \\
\midrule
MART &	\textbf{7.42/11.81}&	\textbf{9.48/16.77}&	0.010 &	 \textcolor{blue}{0.055}\\
MGF &	\textcolor{blue}{7.71/12.07}&	11.30/19.88&	0.019 &  0.058\\
NMRF&	7.86/12.59&	\textcolor{blue}{10.19/18.02}&	\textcolor{blue}{0.004}&	0.066\\
\midrule
Ours & 9.20/15.97 & 11.45/21.02 & \textbf{0.002} & \textbf{0.028}\\
\bottomrule
\end{tabular}
}
\label{tab:sdd_ade}
\end{table}

\begin{figure}[t]
\includegraphics[width=\linewidth]{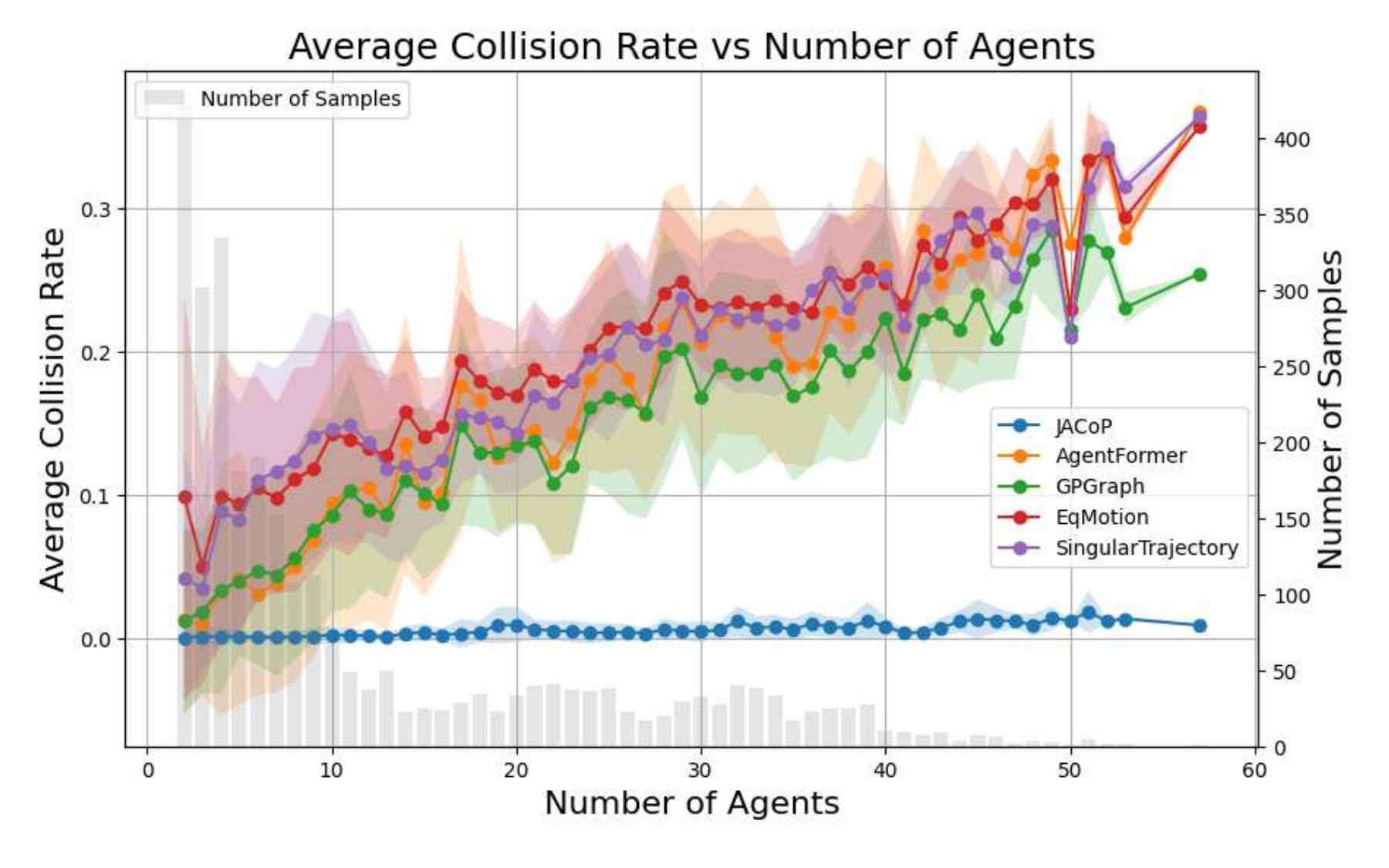}
\vspace{-0.35in}
\caption{A2A collision rate versus number of agents on the ETH-UCY dataset, showing our JACoP model is clearly better at avoiding collisions, especially in crowded scenes.}
\label{fig:rebuttal-col-vs-num-agent}
\vspace{-0.1in}
\end{figure}

\section{Alternative Profiler Choice -- AgentFormer}
\label{sec:supp-alt-baseline}
Our ablation study in Section 4.3 shows that the proposed environmental and social collision filter and the Gibbs sampling module are vital for generating collision-free scene-level predictions. Given this result, we recognize that both the filtering technique and our Joint Prediction Aligner module (Figure 2, right) are generally applicable to other Profiler modules, provided the module is probabilistic or offers a scoring mechanism for trajectory proposals. However, we argue that our light-weight anchor-based Profiler(ACP), which performs matching in the future trajectory's feature space, offers a sufficient and efficient solution. To further verify this hypothesis on the sufficiency of our approach, we experimented with an alternative Profiler choice, using the predictions and probability estimates from AgentFormer \cite{yuan2021agentformer} as inputs to our Joint Prediction Aligner. We benchmark this alternative baseline against the original AgentFormer and our proposed JACoP method.

\begin{figure*}[t]
    \begin{subfigure}[b]{0.196\textwidth}
    \centering
        \includegraphics[width=\linewidth]{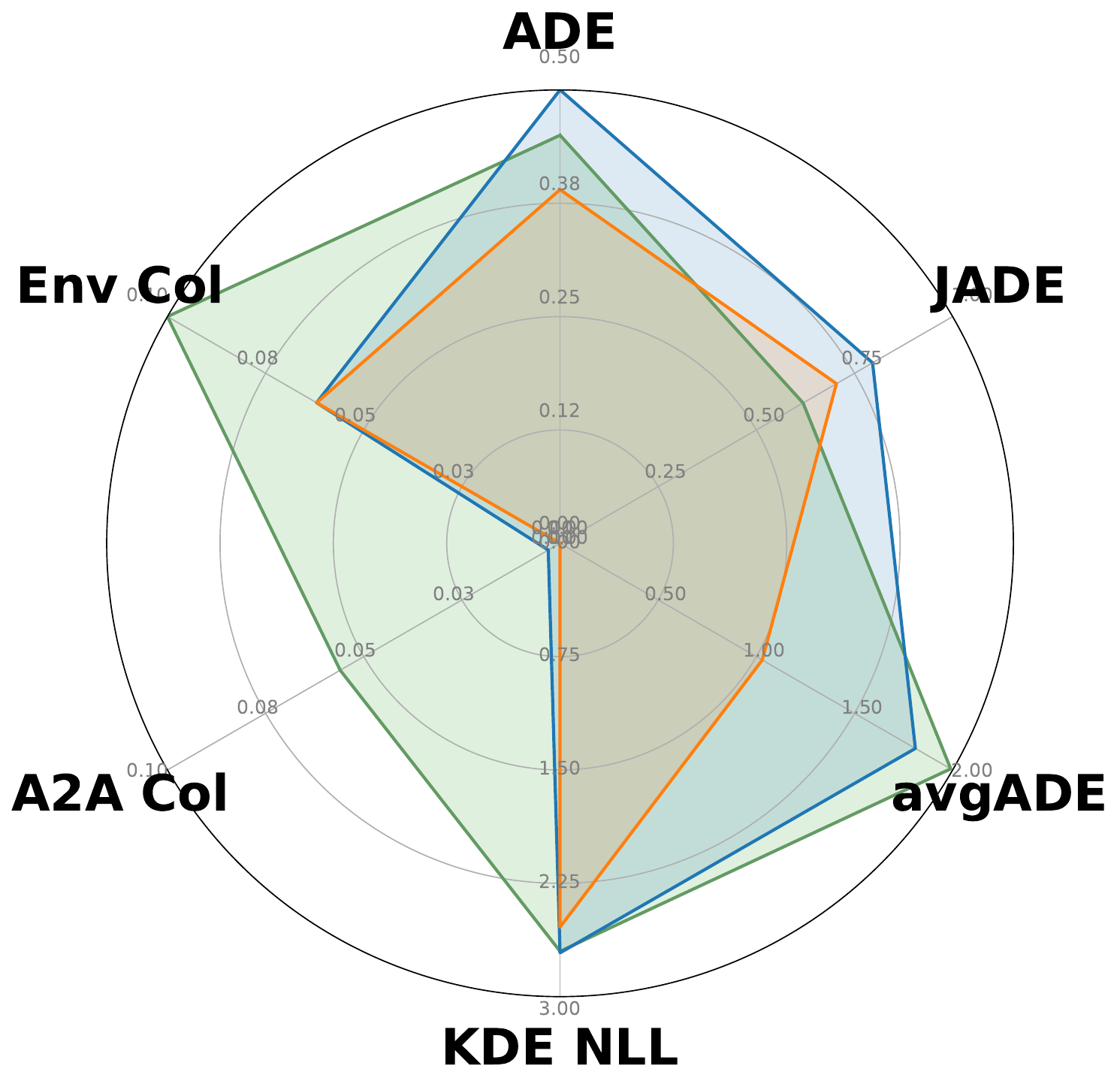}    
        \caption{ETH}
        \end{subfigure}
        \hfill
    \begin{subfigure}[b]{0.196\textwidth}
        \centering
        \includegraphics[width=\linewidth]{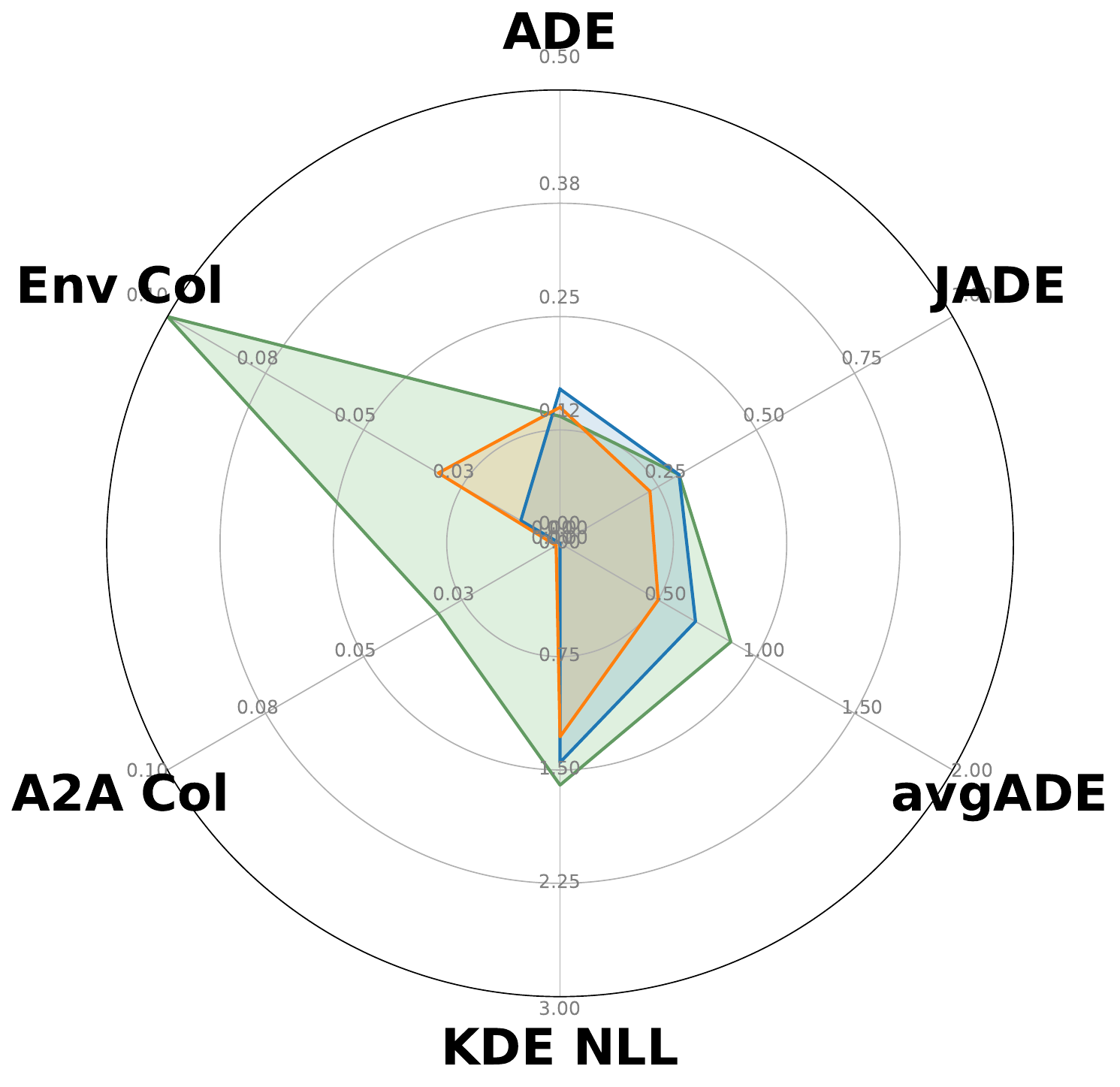}
        \caption{Hotel}
        \end{subfigure}
    \hfill
    \begin{subfigure}[b]{0.196\textwidth}
        \centering
        \includegraphics[width=\linewidth]{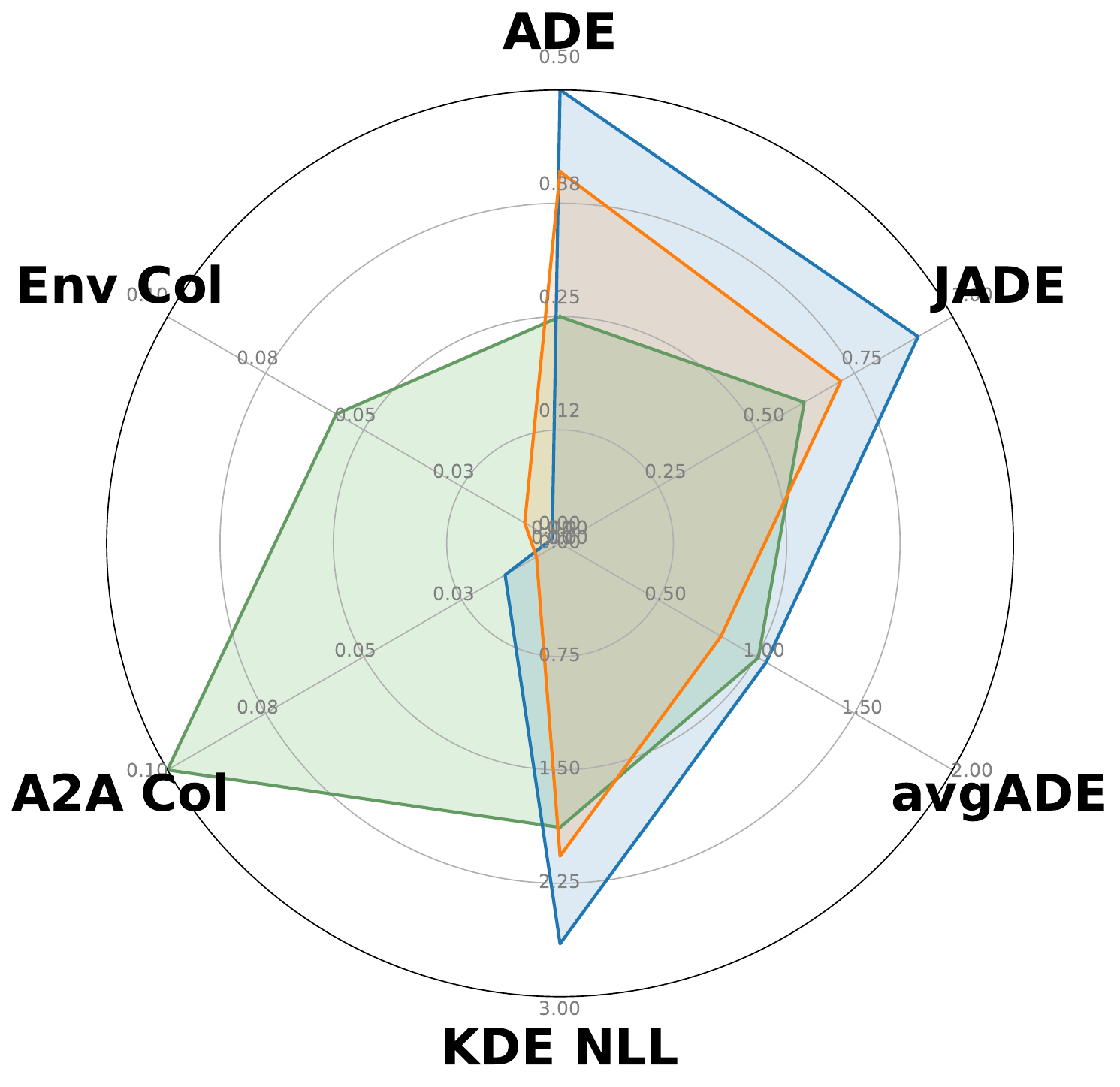}
        \caption{Univ.}
        \end{subfigure}
    \hfill
    \begin{subfigure}[b]{0.196\textwidth}
        \centering
        \includegraphics[width=\linewidth]{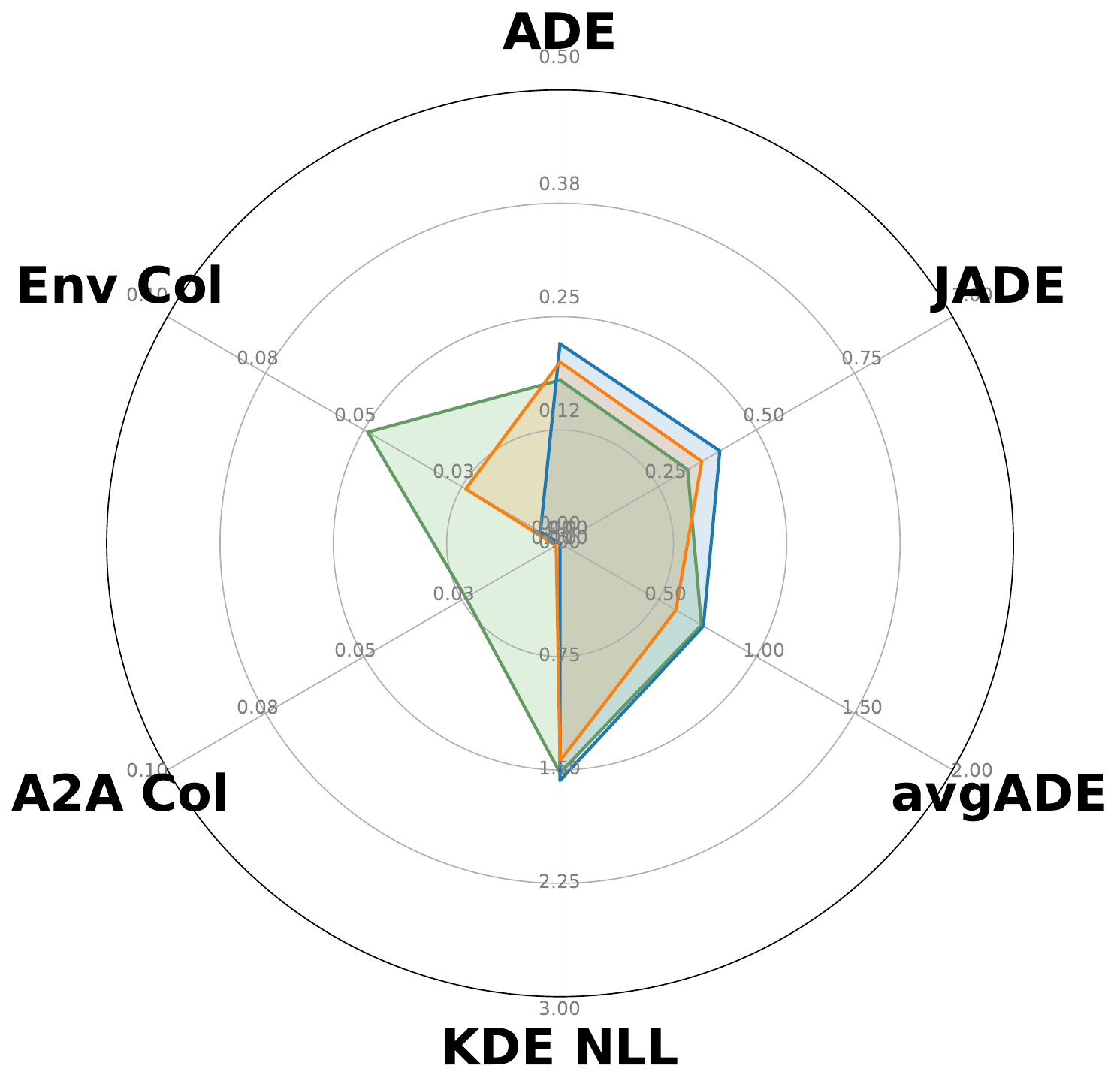}
        \caption{Zara1}
        \end{subfigure}
    \hfill
    \begin{subfigure}[b]{0.196\textwidth}
        \centering
        \includegraphics[width=\linewidth]{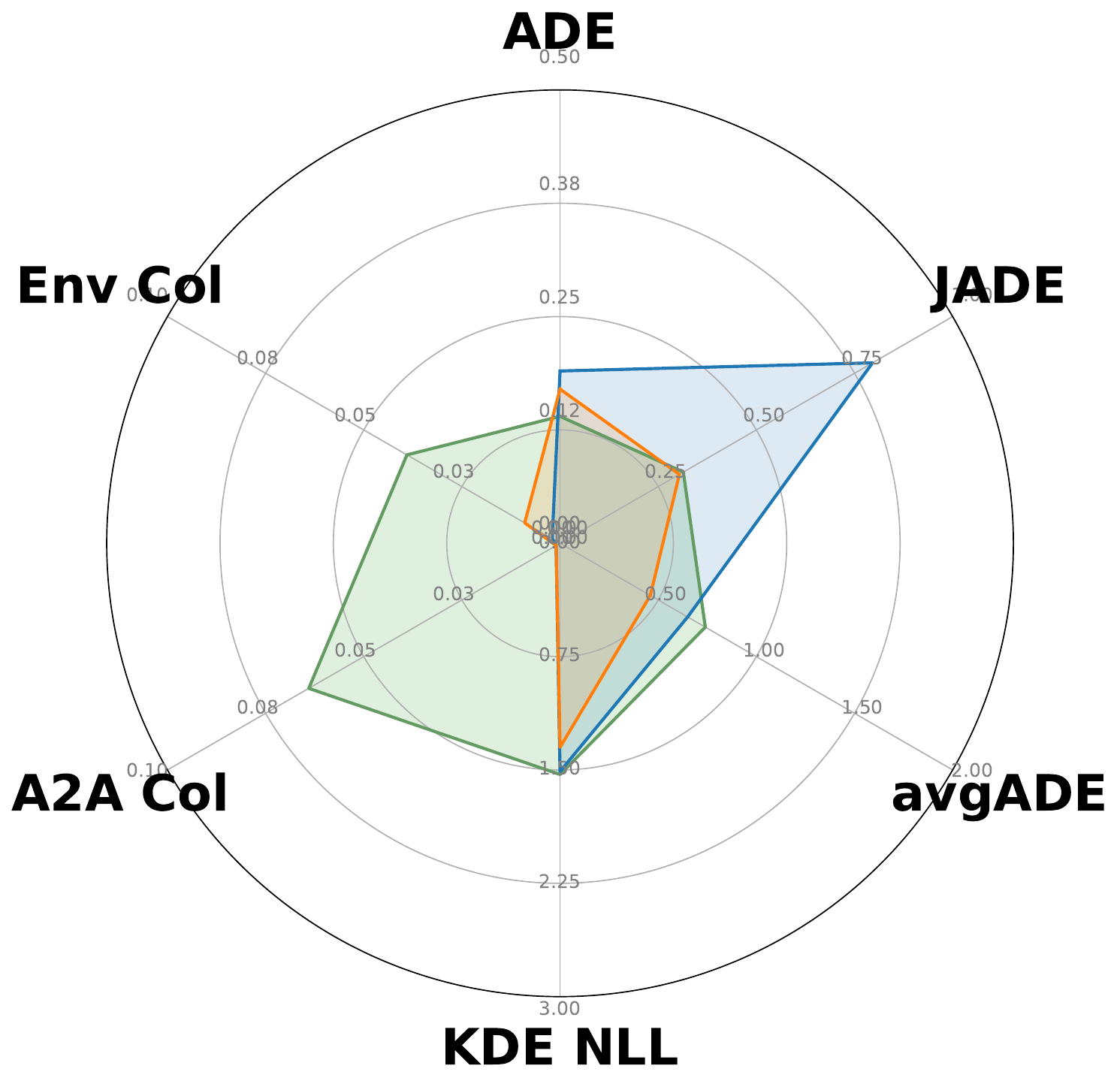}
        \caption{Zara2}
        \end{subfigure}    
    \caption{Radar plot for all metrics among the five ETH-UCY dataset splits comparing between AgentFormer+JACoP Aligner and JACoP. {\color{green}Green: AgentFormer; \color{blue}Blue: AgentFormer Profiler + JACoP Aligner}; {\color{orange}Orange: JACoP}. A smaller overall area of the polygon aligns with better performance across all metrics.}
    \label{fig:radars2}
    \vspace{-0.2in}
\end{figure*}

\Cref{tab:alt_baseline_col}, \Cref{tab:alt_baseline_acc} and \Cref{tab:alt_baseline_avgade_nll} show the performance on the accuracy and feasibility benchmarks, while \Cref{fig:radars2} summarizes the numerical results on the radar charts for better head-to-head comparison. The results concerning environmental and A2A collisions demonstrate that the combination of AgentFormer with our proposed aligner and filtering technique  (shown in blue polygon), significantly reduces the collision rate compared to the original AgentFormer (shown in green polygon). This robustly indicates the utility and generalizability of the JACoP Aligner module.

However, the alternative AgentFormer Profiler fails to outperform our anchor-based Agent-Centric Profiler (ACP) (shown in orange) across the three main accuracy metrics, KDE NLL and the A2A collision rate. Our utilization of environmental and social contexts as part of the query embeddings helps the model select more accurate and suitable prototypes for subsequent alignment and scene prediction generation. While the AgentFormer output shows good performance in the best-case scenario (indicated by the $min_kADE/FDE$ metrics), its inherent lack of environmental awareness renders a majority of its predictions unusable for alignment after the environmental filtering process. The remaining filtered predictions, despite achieving good environmental compliance, lack the necessary accuracy for the alignment process to produce a high-quality final output, which also leads to more unavoidable social collisions. The superior performance of JACoP in the average ADE/FDE and KDE NLL further demonstrate that our proposed ACP module helps the model achieve good average accuracy across all selected samples and generate trajectories that have a tight concentration around the GT future.

\begin{table*}[t]
\centering
\caption{Collision Rate of our JACoP vs. AgentFormer + JACoP Aligner. The best performance is boldfaced and the 2nd place is marked as blue.}
\label{tab:alt_baseline_col}
\resizebox{\textwidth}{!}{
\begin{tabular}{c|ccccc|ccccc}
\toprule
Model & ETH & HOTEL & UNIV & ZARA1 & ZARA2 & ETH & HOTEL & UNIV & ZARA1 & ZARA2 \\
Avg. \# Agents & 2.6 & 3.5 & 25.7 & 3.7 & 6.3 & 2.6 & 3.5 & 25.7 & 3.7 & 6.3 \\
\midrule
Model & \multicolumn{5}{c|}{Environmental Collision} & \multicolumn{5}{c}{A2A Collision}  \\
\midrule
AgentFormer & 0.32&	0.147&	0.057&	0.049& 0.039 & 0.056&	0.031&	0.205&	0.024&	0.064 \\
AgentFormer + JACoP Aligner & \textbf{0.008} & \textbf{0.010}& \textbf{0.000} & \textbf{0.005}& \textbf{0.002} & \textcolor{blue}{0.003} &\textcolor{blue}{0.002}& \textcolor{blue}{0.014}& \textcolor{blue}{0.00}	& \textcolor{blue}{0.002} \\
JACoP & \textcolor{blue}{0.062}& \textcolor{blue}{0.031}& \textcolor{blue}{0.009}& \textcolor{blue}{0.024} & \textcolor{blue}{0.009} & \textbf{0.00}& \textbf{0.001} & \textbf{0.006}& \textbf{0.001} & \textbf{0.001} \\
\bottomrule
\end{tabular}
}
\end{table*}

\begin{table*}[t]
\centering
\caption{Prediction Accuracy Comparison between our JACoP vs. AgentFormer + JACoP Aligner. The best performance is boldfaced and the 2nd place is marked as blue.}
\label{tab:alt_baseline_acc}
\resizebox{\textwidth}{!}{
\begin{tabular}{c|ccccc|ccccc}
\toprule
Model & ETH & HOTEL & UNIV & ZARA1 & ZARA2 & ETH & HOTEL & UNIV & ZARA1 & ZARA2 \\
Avg. \# Agents & 2.6 & 3.5 & 25.7 & 3.7 & 6.3 & 2.6 & 3.5 & 25.7 & 3.7 & 6.3 \\
\midrule
Model & \multicolumn{5}{c|}{$min_kADE/FDE$} & \multicolumn{5}{c}{JADE/JFDE}  \\
\midrule
AgentFormer & \textbf{0.45/0.79}&	\textbf{0.14/0.22}&	\textbf{0.25/0.45}&	\textbf{0.18/0.30}&	\textbf{0.14/0.23} &	\textbf{0.619/1.136}&	\textcolor{blue}{0.303/0.603}&	\textbf{0.622/1.311}&	\textbf{0.325/0.660}&	\textcolor{blue}{0.314/0.663} \\
AgentFormer + JACoP Aligner & \textcolor{blue}{0.52/0.94}  & 0.17/0.29 & 0.62/1.29 & 0.22/0.43 & 0.19/0.37 & 0.796/1.572 & 0.303/0.614  & 0.912/1.942 & 0.407/0.869 & 0.372/0.807 \\
JACoP & 0.59/1.06& \textcolor{blue}{0.15/0.25} & \textcolor{blue}{0.41/0.82}& \textcolor{blue}{0.20/0.39} & \textcolor{blue}{0.17/0.32} & \textcolor{blue}{0.704/1.226} & \textbf{0.229/0.420} & \textcolor{blue}{0.715/1.472} & \textcolor{blue}{0.361/0.724} & \textbf{0.304/0.623}\\
\bottomrule
\end{tabular}
}
\end{table*}

\begin{table*}[t]
\centering
\caption{Average ADE/FDE and KDE-based NLL Benchmark Comparison between our JACoP vs. AgentFormer + JACoP Aligner. The best performance is boldfaced and the 2nd place is marked as blue.}
\label{tab:alt_baseline_avgade_nll}
\resizebox{\textwidth}{!}{
\begin{tabular}{c|ccccc|ccccc}
\toprule
Model & ETH & HOTEL & UNIV & ZARA1 & ZARA2 & ETH & HOTEL & UNIV & ZARA1 & ZARA2 \\
Avg. \# Agents & 2.6 & 3.5 & 25.7 & 3.7 & 6.3 & 2.6 & 3.5 & 25.7 & 3.7 & 6.3 \\
\midrule
Model & \multicolumn{5}{c|}{Average ADE/FDE} & \multicolumn{5}{c}{KDE NLL}  \\
\midrule
AgentFormer & 1.99/4.39	& 0.87/2.01	& \textcolor{blue}{1.01/2.26}	& 0.72/1.63  & 0.74/1.71 & \textcolor{blue}{2.70}	& 1.60	&	\textbf{1.88}&  \textcolor{blue}{1.52} & 1.53\\
AgentFormer + JACoP Aligner & \textcolor{blue}{1.81/3.92}	&\textcolor{blue}{0.69/1.56}	& 1.05/2.26	& \textcolor{blue}{0.73/1.65}  & \textcolor{blue}{0.65/1.48} & 2.71	& \textcolor{blue}{1.45}	& 2.65	&  1.57 & \textcolor{blue}{1.51}\\
JACoP &	\textbf{1.03/1.97}& \textbf{0.50/1.05} & \textbf{0.82/1.70} & \textbf{0.59/1.26}  & \textbf{0.46/0.99} &	\textbf{2.54}&\textbf{1.28}& \textcolor{blue}{2.07}	&\textbf{1.44}  &\textbf{1.35}\\
\bottomrule
\end{tabular}
}
\end{table*}

\begin{figure*}[t]
    \centering
    \includegraphics[width=1\linewidth]{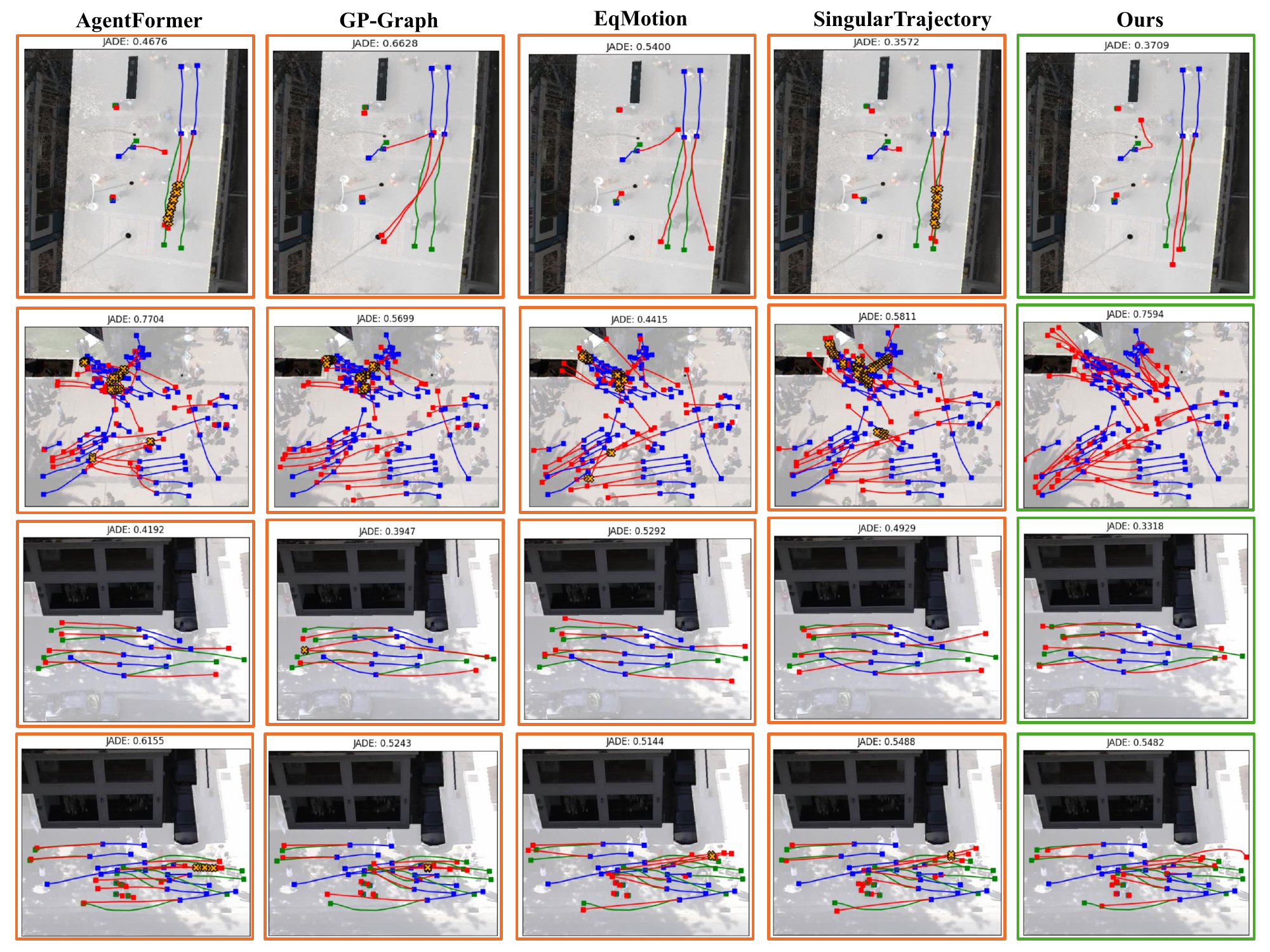}
    \caption{Scene prediction with best JADE performance from Hotel (Row 1), Univ (Row 2), Zara1 (Row 3) and Zara2 (Row 4) \newline \textcolor{Blue}{Blue line:Historical Trajectory}, \textcolor{Green}{Green line: Ground Truth Future Trajectory}, \textcolor{red}{Red line: Predictions}.}
    \label{fig:supp_a2a_col}
\end{figure*}

\begin{figure*}[t]
    \centering
    \includegraphics[width=1\linewidth]{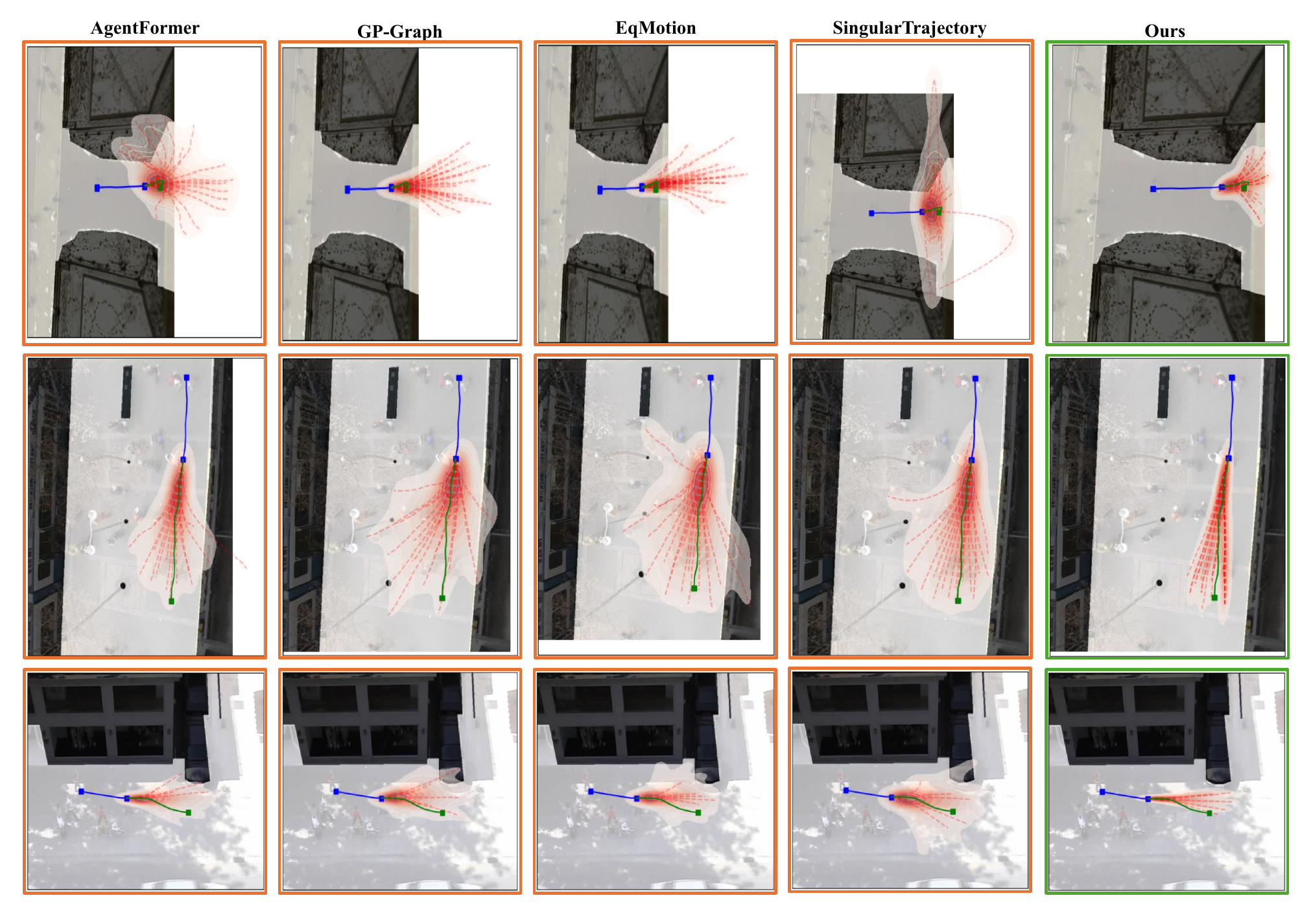}
    \caption{Individual prediction from ETH (Row 1), Hotel (Row 2) and Zara1 (Row 3) \newline \textcolor{Blue}{Blue line:Historical Trajectory}, \textcolor{Green}{Green line: Ground Truth Future Trajectory}, \textcolor{red}{Red Dashed line: Predictions}, \textcolor{red}{Red Shade: Sample Distribution}.}
    \label{fig:supp_env_col}
\end{figure*}

\section{Additional Qualitative Evaluations}
\label{sec:supp-add-qual}
We offer additional visualizations for qualitative analysis. \Cref{fig:supp_a2a_col} presents examples of densely packed scenes, where our JACoP model accurately predicts without social or environmental collisions. For instance, row 2 depicts a crowd navigating dense areas. Compared to other SOTA models, JACoP avoids agent collisions. It also provides accurate scene predictions, evident in the JADE performance for rows 1, 3, and 4.

\Cref{fig:supp_env_col} illustrates individual predictions that demand a deep understanding of environmental constraints. Row 1 features an ETH split example where the model predicts a slowdown as the agent nears the scene's edge. Our model, incorporating environmental context into the query embedding and filtering, generates trajectories that adhere to navigable areas and explores alternatives while respecting constraints. The SingularTrajectory model often fails due to post-prediction corrections, leading to infeasible predictions. Other models, ignoring environmental contexts, tend to overshoot or overlap non-navigable areas. In two other instances, our JACoP models also achieve environmentally compliant predictions, especially in row 2, where our predictions align closely with the ground truth future. 


\begin{algorithm*}
    \caption{Gibbs Sampling for Scene-Level Trajectory Prediction}
    \label{alg:gibbs}
    \begin{algorithmic}[1]
        \Require Number of burn-in steps $B$, number of samples $K$, and a set of agents $\mathcal{I}$.
        \Ensure A set of scene predictions $\mathcal{Y}$.
        
        \State $Y^{(0)} \sim P_{\text{unary}}$ 
        \Comment{Initialize from marginal samples}
        \State $\mathcal{Y} := \text{Empty List}$ 
        
        \For{$\tau = 1$ to $B+K$}

            \For{each agent $i \in \mathcal{I}$}
                \State $Y_i^{(\tau)}\sim P(Y_i|Y^{(\tau-1)}_{\setminus i})$ 
                \Comment{Sample agent $i$'s trajectory conditioned on all others}
            \EndFor
            
            \If {$\tau > B$}
                \State $\text{Append } Y^{(\tau)} \text{ to } \mathcal{Y}$ 
                \Comment{Save the scene prediction after burn-in}
            \EndIf
        \EndFor
        
        \Return $\mathcal{Y}$
    \end{algorithmic}

\end{algorithm*}

\section{Additional Benchmark -- SDD}
\label{sec:supp-sdd}
We added an SDD experiment comparing three SOTA models: MART \cite{lee2024mart}, MGF\cite{chen2024mgf}, and NMRF\cite{fang2025neuralized}, summarized in the radar plot in Figure \ref{fig:sdd-radar}. Our model attains the best social and environmental collision rates—our main focus—with only slightly lower accuracy. We will include full numerical results in the final version.

\begin{table*}[t]
\centering
\caption{Average ADE/FDE and KDE-based NLL Benchmark. The best performance is boldfaced and the 2nd place is marked as blue.}
\vspace{-0.1in}
\label{tab:supp_avgade_nll2}
\resizebox{\textwidth}{!}{
\begin{tabular}{c|ccccc|ccccc}
\toprule
Model & ETH & HOTEL & UNIV & ZARA1 & ZARA2 & ETH & HOTEL & UNIV & ZARA1 & ZARA2 \\
Avg. \# Agents & 2.6 & 3.5 & 25.7 & 3.7 & 6.3 & 2.6 & 3.5 & 25.7 & 3.7 & 6.3 \\
\midrule
Model & \multicolumn{5}{c|}{Average ADE/FDE} & \multicolumn{5}{c}{KDE NLL}  \\
\midrule
AgentFormer & 1.99/4.39	& 0.87/2.01	& 1.01/2.26	& 0.72/1.63  & 0.74/1.71 & 2.71	& 1.35	&	1.72&  \textcolor{blue}{1.30} & 1.29\\
AgentFormer + JACoP Aligner & 1.81/3.92	& 0.69/1.56	& 1.05/2.26	& 0.73/1.65  & 0.65/1.48 & 2.81	& 1.19	&	2.93&  1.38 & 1.29\\
GP-Graph& \textcolor{blue}{1.21/2.51}	& 0.71/1.53	& \textbf{0.91/1.31}	& \textcolor{blue}{0.68/1.50}  & \textcolor{blue}{0.57/1.27} & 2.30	& 1.28	& \textcolor{blue}{1.65}	& 1.32  &\textcolor{blue}{1.19}	\\
EqMotion& 1.42/2.99 & \textcolor{blue}{0.64/1.32}	& 2.30/5.39	& 0.82/1.83	& 0.68/1.54  & \textcolor{blue}{2.15} & \textcolor{blue}{1.10}	& \textbf{1.60}	& 1.40	& 1.22  \\
SingularTrajectory& 1.47/2.77 & 0.89/1.88 & 1.12/2.36 & 0.92/1.96	& 1.02/2.23  &\textbf{1.91} &1.34 & 1.77	&1.46	&1.57  \\
JACoP &	\textbf{1.03/1.97}& \textbf{0.50/1.05} & \textcolor{blue}{0.82/1.70} & \textbf{0.59/1.26}  & \textbf{0.46/0.99} &	2.76&\textbf{1.01}& 2.10	&\textbf{1.24}  &\textbf{1.12}\\
\bottomrule
\end{tabular}
}
\vspace{-0.2in}
\end{table*}

\section{Analysis on Collision Rate vs Number of Scene Agent}
\label{sec:supp-col-vs-num-agent}
We further showcase this robustness via the A2A collision rate (Fig.~\ref{fig:rebuttal-col-vs-num-agent}) across scenes of varying density, where our method decisively outperforms prior SOTA in generating feasible, collision-free predictions.


\end{document}